\documentclass[10pt,journal,compsoc]{IEEEtran}
\pdfoutput=1
\ifCLASSOPTIONcompsoc
  \usepackage[nocompress]{cite}
\else
  \usepackage{cite}
\fi
\ifCLASSINFOpdf
   \usepackage[pdftex]{graphicx}
\else
\fi

\usepackage{cuted}
\usepackage{caption}
\usepackage{float}
\usepackage{bm}
\usepackage{bbold}
\usepackage{amsthm,amssymb}
\usepackage{mathrsfs}
\usepackage{tabularx}
\usepackage{bbding}
\usepackage{ragged2e}
\usepackage{amsmath,amsfonts}
\usepackage{algorithmic}
\usepackage{algorithm}
\makeatletter
\newcommand{\removelatexerror}{\let\@latex@error\@gobble}
\makeatother
\usepackage{array}
\usepackage{textcomp}
\usepackage{stfloats}
\usepackage{url}
\usepackage{verbatim}
\usepackage{graphicx}
\usepackage{cite}
\usepackage{color}
\usepackage{dsfont}
\usepackage{multirow}

\makeatletter

\newcommand{\Rmnum}[1]{\expandafter\@slowromancap\romannumeral #1@}
\makeatother

\usepackage[colorlinks,urlcolor= black,linkcolor=black,citecolor=black]{hyperref}

\ifCLASSOPTIONcompsoc
 \usepackage[caption=false,font=footnotesize,labelfont=sf,textfont=sf]{subfig}
\else
 \usepackage[caption=false,font=footnotesize]{subfig}
\fi

\begin{document}

\title{Pyramid Self-attention Polymerization Learning for Semi-supervised Skeleton-based Action Recognition}

\author{Binqian~Xu, Xiangbo~Shu,~\IEEEmembership{Senior Member, IEEE},
\IEEEcompsocitemizethanks{\IEEEcompsocthanksitem B. Xu, and X. Shu are with the School of Computer Science and Engineering, Nanjing
			University of Science and Technology, Nanjing 210094, China. E-mail: xubinq11@gmail.com, shuxb@njust.edu.cn. Corresponding author: Xiangbo Shu.
}
}

\markboth{Submission~of~IEEE~TRANSACTIONS~ON~PATTERN~ANALYSIS~AND~MACHINE~INTELLIGENCE, 2022}%
{Submission~of~IEEE~TRANSACTIONS~ON~PATTERN~ANALYSIS~AND~MACHINE~INTELLIGENCE, 2022}

\IEEEtitleabstractindextext{%
\begin{abstract}
\justifying
Most semi-supervised skeleton-based action recognition approaches aim to learn the skeleton action representations only at the joint level, but neglect the crucial motion characteristics at the coarser-grained body (e.g., limb, trunk) level that provide rich additional semantic information, though the number of labeled data is limited. In this work, we propose a novel Pyramid Self-attention Polymerization Learning (dubbed as PSP Learning) framework to jointly learn body-level, part-level, and joint-level action representations of joint and motion data containing abundant and complementary semantic information via contrastive learning covering coarse-to-fine granularity. Specifically, to complement semantic information from coarse to fine granularity in skeleton actions, we design a new Pyramid Polymerizing Attention (PPA) mechanism that firstly calculates the body-level attention map, part-level attention map, and joint-level attention map, as well as polymerizes these attention maps in a level-by-level way (i.e., from body level to part level, and further to joint level). Moreover, we present a new Coarse-to-fine Contrastive Loss (CCL) including body-level contrast loss, part-level contrast loss, and joint-level contrast loss to jointly measure the similarity between the body/part/joint-level contrasting features of joint and motion data. Finally, extensive experiments are conducted on the NTU RGB+D and North-Western UCLA datasets to demonstrate the competitive performance of the proposed PSP Learning in the semi-supervised skeleton-based action recognition task. The source codes of PSP Learning are publicly available at \url{https://github.com/1xbq1/PSP-Learning}.
\end{abstract}

\begin{IEEEkeywords}
Action recognition, Skeleton, Semi-supervised, Contrastive learning, Self-attention.
\end{IEEEkeywords}
}

\maketitle

\IEEEdisplaynontitleabstractindextext

\IEEEpeerreviewmaketitle

\IEEEraisesectionheading{\section{Introduction}\label{sec:introduction}}
\IEEEPARstart{H}{uman} action recognition is an attractive topic in the area of computer vision, due to its significant role in various latent applications, e.g., human-computer interaction, video surveillance, video management, etc~\cite{weinland2011survey,wang2013action,simonyan2014two,tran2015learning,carreira2017quo,shu2019hierarchical,shu2020host,tang2019coherence}. Recently, skeleton-based action recognition has been studied extensively, this is because: i) Skeleton data is easier to be acquired by many depth sensors or pose estimation algorithms~\cite{zhang2012microsoft,cao2017realtime}; ii) Skeleton data usually represented by the coordinate information is robust for dynamic circumstances, human body scales, and viewpoint variations~\cite{kim2017interpretable,li2018independently,shu2021spatiotemporal}; iii) Skeleton data has more advantages in calculation and storage, compared with the scale of RGB data~\cite{li2018co,yan2017skeleton}. 

\begin{figure}[!t]
    \centering
    \includegraphics[width=3.4in]{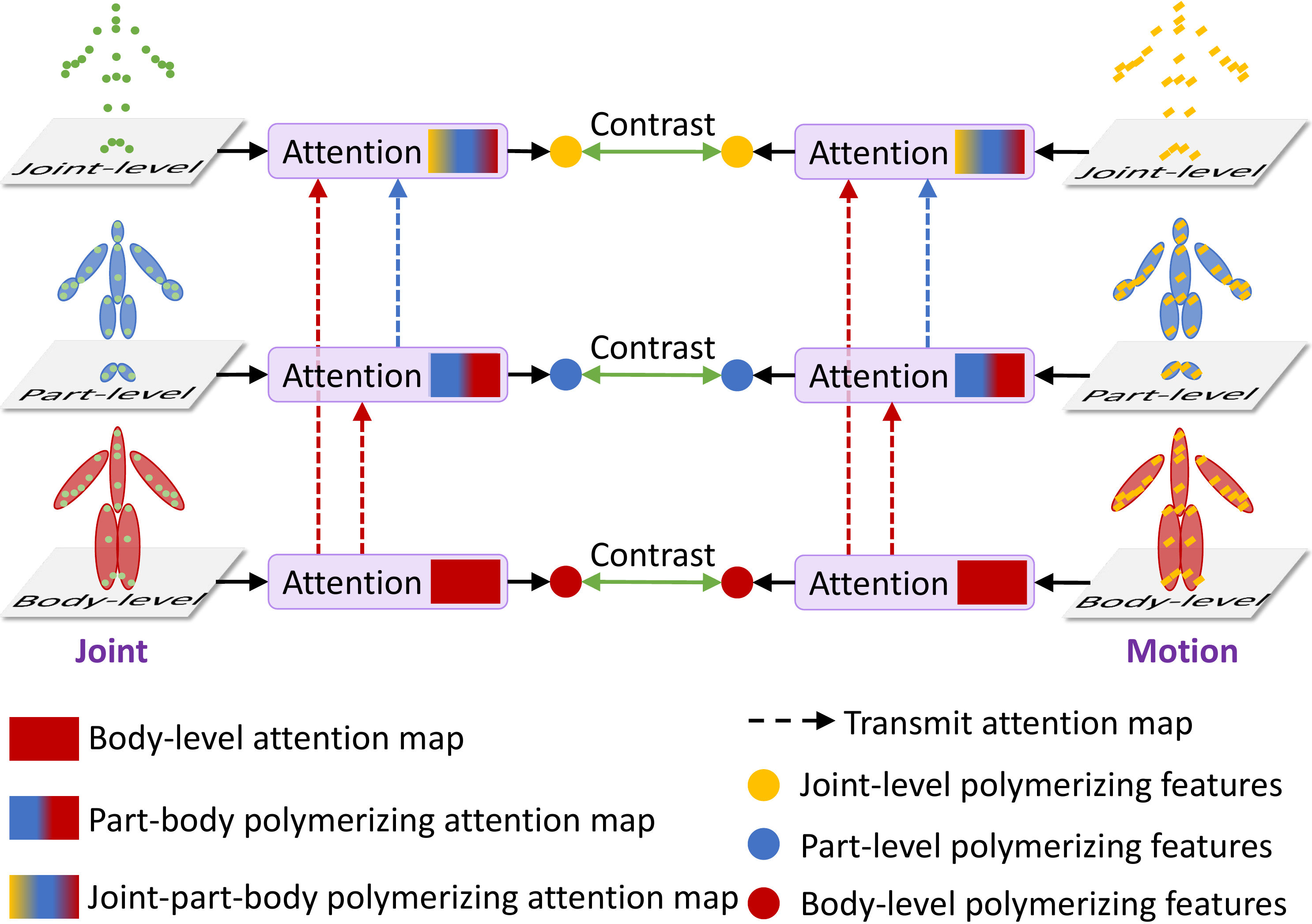}
    \caption{The main idea of this work. The skeleton can be transformed into a coarse-to-fine skeleton pyramid structure (including body level, part level, and joint level), wherein the corresponding body-level, part-level, and joint-level features of joints and motions are obtained. Then, we polymerize the coarse-to-fine self-attentions to obtain body-level, part-level, and joint-level polymerizing features of joints and motions, which are subsequently contrasted in contrastive learning covering coarse-to-fine granularity.}
    \label{idea}
\end{figure}

Deep learning-based methods on the skeleton-based action recognition task have shown outstanding advantages and evolved into various networks in the past ten years~\cite{du2015hierarchical,liu2017skeleton,li2017skeleton,yan2018spatial,li2019actional,shi2020decoupled,wang2021iip}. In particular, Convolutional Neural Network (CNN) and Recurrent Neural Network (RNN) based methods mostly treat the skeleton features as pseudo-images and temporal-related sequences, according to their intrinsic structure. After further investigating the natural structure of the skeleton data, Graph Convolutional Network (GCN) based methods are introduced to construct skeleton data as a graph topology. Recently, some Transformer-based methods perform better for learning features by capturing the relationships among all joints through self-attention~\cite{shi2020decoupled,wang2021iip}. However, most of the above methods adopt the fully supervised training way that relies on a large number of expensive labeled data. Therefore, to decrease the demand for labeled data, various semi-supervised learning methods are presented by learning features from both unlabeled skeleton data and labeled skeleton data~\cite{si2020adversarial, li2020sparse,tu2022joint,lin2020ms2l,su2020predict,xu2021prototypical,li20213d,guo2021contrastive,thoker2021skeleton}.

\begin{figure*}[!t]
    \centering
    \includegraphics[width=7.1in]{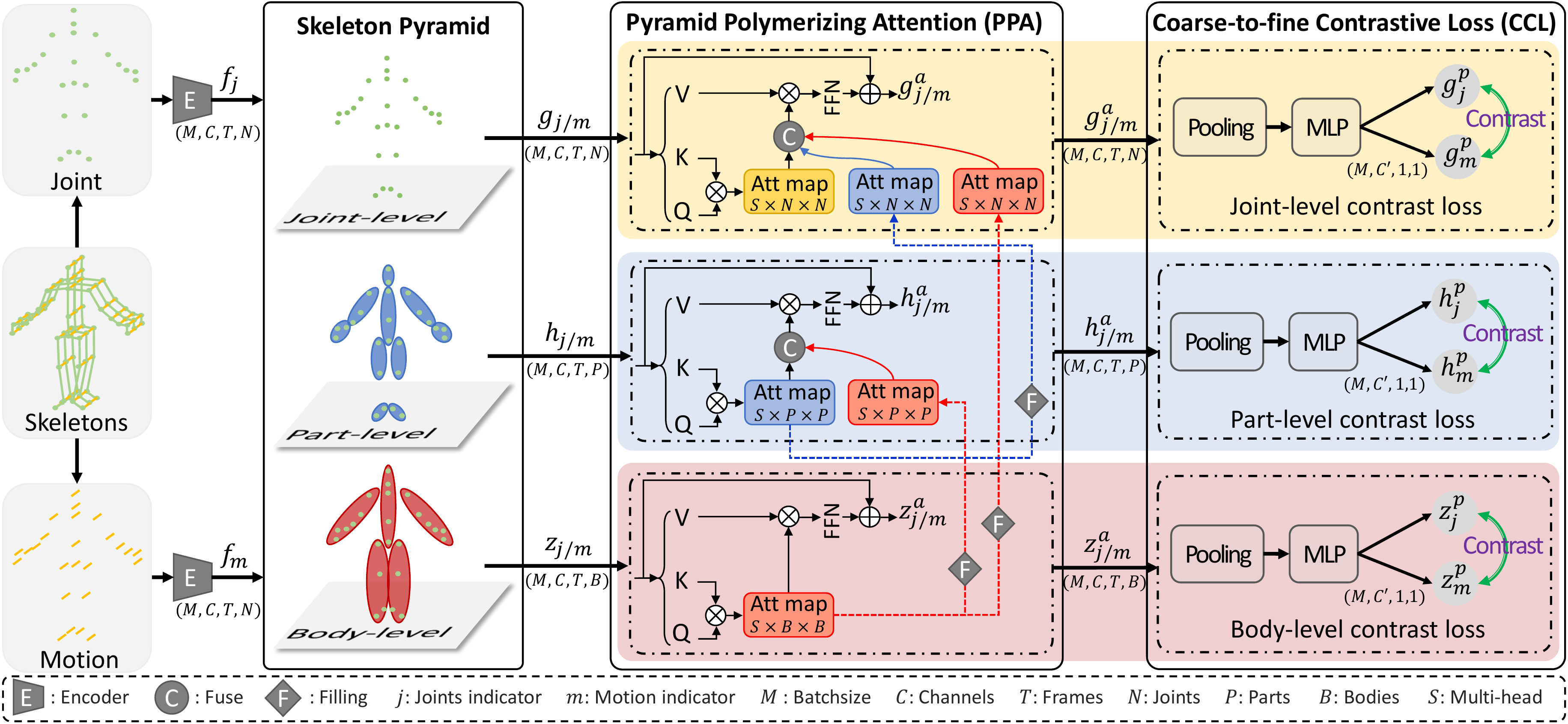}
    \caption{
    {The overall framework of Pyramid Self-attention Polymerization Learning (PSP Learning). PSP Learning mainly consists of Encoder, Skeleton Pyramid (including body level, part level, and joint level), Pyramid Polymerizing Attention (PPA), and Coarse-to-fine Contrastive Loss (CCL) (consisting of body-level contrast loss, part-level contrast loss, and joint-level contrast loss). The joint and motion data in skeleton sequences are fed into the Encoder under the Skeleton Pyramid structure to obtain the body-level, part-level, and joint-level features of joint and motion. Then, PPA polymerizes the body-level attention map, part-level attention map, and joint-level attention map from body-to-part-to-joint level for further obtaining body-, part-, and joint-level polymerizing features, which contain abundant and complementary semantic information from coarse to fine granularity. At last, we present a new Coarse-to-fine Contrastive Loss (CCL) to measure the similarity between the body/part/joint-level contrasting features of joint and motion via the body/part/joint-level contrast loss covering coarse-to-fine granularity.}}
    \label{framework}
\end{figure*}

Previous semi-supervised learning-based action recognition methods only focus on the representation learning of the joint-level skeleton, ignoring some additional semantic information at the coarser-grained level (e.g., hand, lower limb). For example, the action ``hand waving" regards the hand as the basic unit of movement, and the action ``walking towards each other" is to alternately move the lower limbs, where both of part-level ``hand" and body-level ``lower limb" are regarded as the coarser granularities directly reflecting the semantic information compared with fine granularity, e.g., bone joint. As such, learning abundant semantic information from such coarse-to-fine granularity is meaningful and reasonable. Preliminarily, we systematically transform the human skeleton to a coarse-to-fine skeleton pyramid structure including body level, part level, and joint level, inspired by prior knowledge of human body structure.
Here, we cannot omit the fact that the body level, part level, and joint level information are not independent. Accordingly, one solution is that we can leverage the self-attention mechanism to polymerize semantic information from the body-to-part-to-joint level (i.e., from body level to part level, and further to joint level). Finally, we contrast multi-level features via contrastive learning covering all granularities. Figure~\ref{idea} shows the main idea of this work.

Based on the above idea, we formally propose a novel Pyramid Self-attention Polymerization Learning (PSP Learning) framework to learn body-level, part-level, and joint-level skeleton action representations containing complementary semantic information via contrastive learning covering coarse-to-fine granularity. Specifically, to complement coarse-to-fine semantic information, we introduce a new Pyramid Polymerizing Attention (PPA) mechanism to calculate and polymerize the body-level, part-level, and joint-level attention map from the body-to-part-to-joint level, as well as produce the body-level, part-level, and joint-level polymerizing features. Subsequently, we present a new Coarse-to-fine Contrastive Loss (CCL) consisting of body-level contrast loss, part-level contrast loss, and joint-level contrast loss to measure the similarity of body/part/joint-level contrasting features of joint and motion.

Figure~\ref{framework} illustrates the main framework of the proposed PSP Learning. It mainly consists of Encoder~\cite{shi2020decoupled}, Skeleton Pyramid (including body level, part level, and joint level), Pyramid Polymerizing Attention (PPA), and Coarse-to-fine Contrastive Loss (CCL) (consisting of body-level contrast loss, part-level contrast loss, and joint-level contrast loss). First, raw skeletons are converted to joint data and motion data by data pre-processing. Subsequently, all pre-processed data are input into the Encoder for extracting joint and motion features, which are further transformed to body-level, part-level, and joint-level ones guided by Skeleton Pyramid. Second, PPA transforms all body-, part-, and joint-level features to body-level, part-level, and joint-level polymerizing features by polymerizing the body-level, part-level, and joint-level attention maps from the body-to-part-to-joint level. Third, body-, part-, and joint-level polymerizing features are fed into CCL, for correspondingly measuring their similarity between joint modality and motion modality. 
Finally, we show the competitive performance of the proposed PSP Learning in terms of semi-supervised skeleton-based action recognition tasks by conducting extensive experiments on both public NTU RGB+D and North-Western UCLA datasets.

On the whole, we summarize the main contributions of this work as follows,
\begin{itemize}
        \item We propose a novel Pyramid Self-attention Polymerization Learning (PSP Learning) framework to jointly learn
body-, part- and joint-level skeleton action representations reflecting abundant and complementary semantic information via contrastive
learning covering coarse-to-fine levels.
        \item We present a new Pyramid Polymerizing Attention (PPA) mechanism to complement semantic information from coarse to fine granularity in skeleton actions by polymerizing the body-level attention map, part-level attention map, and joint-level attention map from the body-to-part-to-joint level.
        \item We design a new Coarse-to-fine Contrastive Loss (CCL) consisting of body-level contrast loss, part-level contrast loss, and joint-level contrast loss to measure the similarity between the body/part/joint-level contrasting features of joint and motion data covering coarse-to-fine granularity.
	\end{itemize}

The rest parts of this paper are organized as follows. Section~\ref{related work} introduces related works on supervised/semi-supervised skeleton-based action recognition, contrastive learning, and self-attention. Section~\ref{methodology} illustrates the proposed PSP Learning in detail for semi-supervised skeleton-based action recognition. Experiments results and analysis are presented in Section~\ref{experiments}, followed by conclusions in Section~\ref{conclusion}.

\section{Related Work}
\label{related work}
In this section, we mainly survey some articles related to supervised skeleton-based action recognition, and semi-supervised skeleton-based action recognition, as well as introduce some preliminary methods of contrastive learning and self-attention. 
\subsection{Supervised Skeleton-based Action Recognition}
In supervised learning, the action representation in skeleton data has been shifted from the hand-crafted manner~\cite{wang2013learning,vemulapalli2014human} to the deep learning manner~\cite{du2015hierarchical,veeriah2015differential,zhang2017view,kim2017interpretable,du2015skeleton,liu2017enhanced,yan2018spatial,shi2019two}, due to the latter's powerful representation ability.

As one of the deep learning models, CNNs have shown outstanding performance in various tasks related to 2D image content analysis and understanding~\cite{song2020stronger,chen2021channel,shi2020decoupled,wang2021iip}. In CNN-based skeleton-based action recognition methods, they generally treat the spatial-temporal skeleton data as pseudo-images~\cite{wang2016action,hou2016skeleton,zhang2019view}. For example, Wang et al.~\cite{wang2016action} encoded 3D skeleton sequence as multiple 2D images, which are fed into CNNs to get the discriminative features. Compared with CNNs, RNNs are more beneficial for learning the dynamic dependence of sequence data, so they are usually used to model the temporal context dynamic of skeleton sequences~\cite{du2015hierarchical,veeriah2015differential,zhang2017view}. For example, Du et al.~\cite{du2015hierarchical} divided the human skeleton into five body parts, and then fed these body parts into multiple RNNs to obtain a time-accumulated representation. Different from RNNs and CNNs, GCNs are more conducive to learning structural skeleton data by exploring the inter-relationships among body joints~\cite{yan2018spatial,tang2018deep,li2019actional,shi2019two,song2020stronger,liu2020disentangling,shi2021adasgn,chi2022infogcn}. For example, Yan et al.~\cite{yan2018spatial} proposed a spatial-temporal graph convolution module to model dynamic human skeleton data by regarding joints, human-intrinsic and continuous-frame connections as nodes, spatial and temporal edges. To embed the physical significant knowledge into action representations, Chi et al.~\cite{chi2022infogcn} introduced an informative bottleneck between input and learned multi-modal representations, and also designed a self-attention graph convolution for inferring the context-dependent topology. Recently, Transformer has shown great promise for the processing and modeling of sequence data, so many Transformer-based methods have emerged to model the spatiotemporal information of skeleton sequences~\cite{shi2020decoupled,plizzari2021spatial,zhang2021stst,wang2021iip,liu2022graph,qiu2022spatio}. For example, to alleviate the joint-level noise and save on computation and storage overhead, Wang et al.~\cite{wang2021iip} employed Transformer to capture intra- and inter-part dependencies on part-level skeleton data and masked data. Inspired by self-attention, Shi et al.~\cite{shi2020decoupled} presented a spatial-temporal attention decoupling model with decoupled position encoding to learn action representations for skeleton sequences. Although the above methods have achieved outstanding performance, they are all trained in a fully supervised way and rely on a large amount of labeled skeleton data.

\subsection{Semi-Supervised Skeleton-based Action Recognition}
In semi-supervised learning, the informative representations are learned from both unlabeled data and labeled data~\cite{chapelle2009semi}. Since the amount of unlabeled data is much larger than that of labeled data sometimes, exploring meaningful representations from unlabeled data is the key to improving performance of corresponding semi-supervised learning tasks. Specifically, skeleton representations containing more discriminative dynamic information can be also gained from unlabeled data, which is beneficial to the semi-supervised action recognition tasks~\cite{si2020adversarial,li2020iterate,li2020sparse,tu2022joint,xu2022x}. 
For example, Si et al.~\cite{si2020adversarial} proposed a neighborhood consistency self-supervised learning with an adversarial regularization to align feature distributions for generating the final meaningful semi-supervised representations. In order to use as few labels as possible, Li et al.~\cite{li2020iterate} presented an active learning strategy to select the most meaningful labeled data based on the skeleton reconstruction of the encoder-decoder. Except for learning joint- and bone-modal information separately, Tu et al.~\cite{tu2022joint} presented a graph convolution for joint and bone information fusion by transferring motion information across the joint stream and bone stream. Recently, Xu et al.~\cite{xu2022x} proposed an X-invariant contrastive augmentation and representation learning framework to learn augmentations and representations of skeleton sequences via contrastive learning. In this work, we consider learning the multi-granularity representations via coarse-to-fine contrastive learning at the body level, part level, and joint level.

\subsection{Contrastive Learning}
Contrastive learning has made breakthroughs in recent years, and attracted widespread attention in various fields, due to its outstanding performance in representation learning~\cite{he2020momentum,chen2020simple,tian2020makes,grill2020bootstrap,caron2020unsupervised, gao2021contrastive,rao2021augmented}. As one of the classical works, He et al.~\cite{he2020momentum} proposed to use a queue to store samples and introduced the Encoder with momentum update for constructing a large and consistent dynamic dictionary for contrastive learning. Subsequently, Chen et al.~\cite{chen2020simple} proposed to contrast among large batch sizes of the samples via various augmentation operations to further boost the representations.

In the field of skeleton-based action recognition, there are some competitive methods based on contrastive learning, focusing on various data augmentation strategies~\cite{gao2021contrastive,rao2021augmented,guo2021contrastive,zhan2021spatial,moliner2022bootstrapped}, or different contrastive pretext tasks~\cite{lin2020ms2l,xu2021prototypical,wang2021contrast,thoker2021skeleton,li20213d,zhang2022contrastive,mao2022cmd}. For example, Gao et al.~\cite{gao2021contrastive} proposed the combined augmentations of viewpoint and distance transforms for contrastive learning. Meanwhile, Rao et al.~\cite{rao2021augmented} tested the effect of various augmentations (including rotation, shear, reverse, gaussian noise, gaussian blur, joint mask, and channel mask) of skeleton data in contrastive learning. To alleviate the over-fitting problem in single-task learning and insufficient generalization of the learned features, Lin et al.~\cite{lin2020ms2l} explored to learn rich action representations by three different tasks (i.e., predicting motion, recognizing jigsaw puzzle, and contrasting). 
For obtaining more reliable information on contrast pairs, Li et al.~\cite{li20213d} proposed to learn the cross-view semantic representations based on the consistency of semantic information among different views in the pretext task of contrastive learning. In this work, we learn the cross-modality representations based on the consistency of semantic information among different granularities, i.e., body level, part level, and joint level.

\subsection{Self-Attention Mechanism}
The self-attention mechanism is a variant of the attention mechanism that focuses on the dependencies within data~\cite{vaswani2017attention,khan2021transformers}. Specifically, the query $Q$, key $K$, and value $V$ with dimension $d$ are firstly calculated from the input via three learnable weight matrices. Then the dot product of the query and key is divided by $\sqrt{d}$ for gradient stabilization.
Finally, the output is obtained by multiplying the normalized dot product result and value. The overall process can be expressed as follows,
\begin{equation}
    \text{Attention}(Q,K,V) = \text{softmax}(\frac{QK^\top}{\sqrt{d}})V
\end{equation}
In addition, the self-attention module focusing on important information is the key component of Transformer~\cite{vaswani2017attention}. More details and applications about Transformer can be found in recent survey articles~\cite{khan2021transformers,han2022survey,liu2021survey}. Evolved from the conventional self-attention mechanism, we present a new Pyramid Polymerizing Attention (PPA) mechanism to polymerize self-attentions from coarse to fine granularity for capturing more complement semantic information of skeleton actions.

\section{Methodology}
\label{methodology}
\subsection{Overview of PSP Learning}
\begin{table}[!t]
    \renewcommand{\arraystretch}{1.4}
    \centering
    \caption{Definition of some notations.}
    \begin{tabular}{l|l|l}
        \hline
        Notation & Definition & Dimension \\ \hline
        $f_j$ & Joint features & $\mathbb{R}^{M\times C\times T\times N}$ \\
        $f_m$ & Motion features & $\mathbb{R}^{M\times C\times T\times N}$ \\
        $g_j$ & Joint-level joint features & $\mathbb{R}^{M\times C\times T\times N}$ \\
        $g_m$ & Joint-level motion features & $\mathbb{R}^{M\times C\times T\times N}$ \\
        $h_j$ & Part-level joint features & $\mathbb{R}^{M\times C\times T\times P}$ \\
        $h_m$ & Part-level motion features & $\mathbb{R}^{M\times C\times T\times P}$ \\
        $z_j$ & Body-level joint features & $\mathbb{R}^{M\times C\times T\times B}$ \\
        $z_m$ & Body-level motion features & $\mathbb{R}^{M\times C\times T\times B}$ \\
        $g^a_j$ & Joint-level polymerizing joint features & $\mathbb{R}^{M\times C\times T\times N}$ \\
        $g^a_m$ & Joint-level polymerizing motion features & $\mathbb{R}^{M\times C\times T\times N}$ \\
        $h^a_j$ & Part-level polymerizing joint features &  $\mathbb{R}^{M\times C\times T\times P}$ \\
        $h^a_m$ & Part-level polymerizing motion features & $\mathbb{R}^{M\times C\times T\times P}$ \\
        $z^a_j$ & Body-level polymerizing joint features & $\mathbb{R}^{M\times C\times T\times B}$ \\
        $z^a_m$ & Body-level polymerizing motion features & $\mathbb{R}^{M\times C\times T\times B}$ \\
        $g^p_j$ & Joint-level contrasting joint features & $\mathbb{R}^{M\times C'\times 1\times 1}$ \\
        $g^p_m$ & Joint-level contrasting motion features & $\mathbb{R}^{M\times C'\times 1\times 1}$ \\
        $h^p_j$ & Part-level contrasting joint features & $\mathbb{R}^{M\times C'\times 1\times 1}$ \\
        $h^p_m$ & Part-level contrasting motion features & $\mathbb{R}^{M\times C'\times 1\times 1}$ \\
        $z^p_j$ & Body-level contrasting joint features & $\mathbb{R}^{M\times C'\times 1\times 1}$ \\
        $z^p_m$ & Body-level contrasting motion features & $\mathbb{R}^{M\times C'\times 1\times 1}$ \\
        \hline
    \end{tabular}
    \label{tab_notations}
\end{table}
The overall framework of the proposed PSP Learning is illustrated in Figure~\ref{framework}. Given the input skeleton sequences, the different-modality data (i.e., joint data and motion data) generated from them describe the same action category in different ways. Specifically, the joint data denoted by $\mathcal{X}_j$ represents the coordinate position of each joint point of the human body. The motion data denoted by $\mathcal{X}_m$ accordingly supplements the motion difference information of the same joint points in the next frame and the previous frame. Then $\mathcal{X}_j$ and $\mathcal{X}_m$ are fed into the Encoder~\cite{shi2020decoupled} that outputs the corresponding joint features $f_j$ and motion features $f_m$, where $f_j$, $f_m \in \mathbb{R}^{M\times C\times T\times N}$, $M$, $C$, $T$, and $N$ are the batch size, number of channels, frames, and joint-level nodes, respectively. For convenience, some important notations are defined in Table~\ref{tab_notations}. 

From the perspective of the basic unit of the human body performing actions, different levels of movement units in the same action also contain differentiated information, which is able to enrich the semantic information of actions. Therefore, we introduce a Skeleton Pyramid structure to transform joint/motion features $f_{j/m}$ produced by the Encoder from coarse granularity (e.g., body level) to fine granularity (e.g., joint level), based on the biological structure of the human body. To be specific, $f_{j/m}$ is transformed to body-level features  $z_{j/m}$ , part-level features $h_{j/m}$ and joint-level features $g_{j/m}$ based on regular movement patterns of human, where $z_{j/m} \in \mathbb{R}^{M\times C\times T\times B}$, $h_{j/m} \in \mathbb{R}^{M\times C\times T\times P}$, $g_{j/m} \in \mathbb{R}^{M\times C\times T\times N}$, $B$ and $P$ denote the number of body-level nodes and part-level nodes. Since there is a certain correlation between different levels, we present Pyramid Polymerizing Attention (PPA) to polymerize body-to-part-to-joint level information in a layer-by-layer way. In PPA, the body-level attention map $\mathcal{A}_z \in \mathbb{R}^{M\times S\times B\times B}$, part-level attention map $\mathcal{A}_h \in \mathbb{R}^{M\times S\times P\times P}$, and joint-level attention map $\mathcal{A}_g \in \mathbb{R}^{M\times S\times N\times N}$ are calculated by self-attention mechanism. Then, the body-level polymerizing features $z_{j/m}^a$ , part-level polymerizing features $h_{j/m}^a$ and joint-level polymerizing features $g_{j/m}^a$ are obtained from $z_{j/m}$ , $h_{j/m}$ and $g_{j/m}$ by polymerizing attention maps from body-to-part-to-joint level, where $z_{j/m}^a \in \mathbb{R}^{M\times C\times T\times B}$, $h_{j/m}^a \in \mathbb{R}^{M\times C\times T\times P}$, and $g_{j/m}^a \in \mathbb{R}^{M\times C\times T\times N}$. Finally, we also design a Coarse-to-fine Contrastive Loss (CCL) to contrast the similarity of body/part/joint-level features between joint modality and motion modality for promoting comprehensive action representation learning. In CCL, body-level contrasting features $z_{j/m}^p$, part-level contrasting features $h_{j/m}^p$, and joint-level contrasting features $g_{j/m}^p$ , converted from $z_{j/m}^a$ , $h_{j/m}^a$ and $g_{j/m}^a$ via Pooling and MLP, are contrasted between joints and motions by body-level contrast loss, part-level contrast loss, and joint-level contrast loss, where $z_{j/m}^p$ , $h_{j/m}^p$ , $g_{j/m}^p \in \mathbb{R}^{M\times C'\times 1\times 1}$. The whole semi-supervised training process of the proposed PSP Learning can be briefly summarized as training jointly on both labeled and unlabeled data, namely using CCL with unlabeled data to train the Encoder and PPA, while using a recognition loss (e.g., cross-entropy loss) to train the Encoder with a softmax layer.

\subsection{Skeleton Pyramid}
\begin{figure}
    \centering
    \includegraphics[width=3.4in]{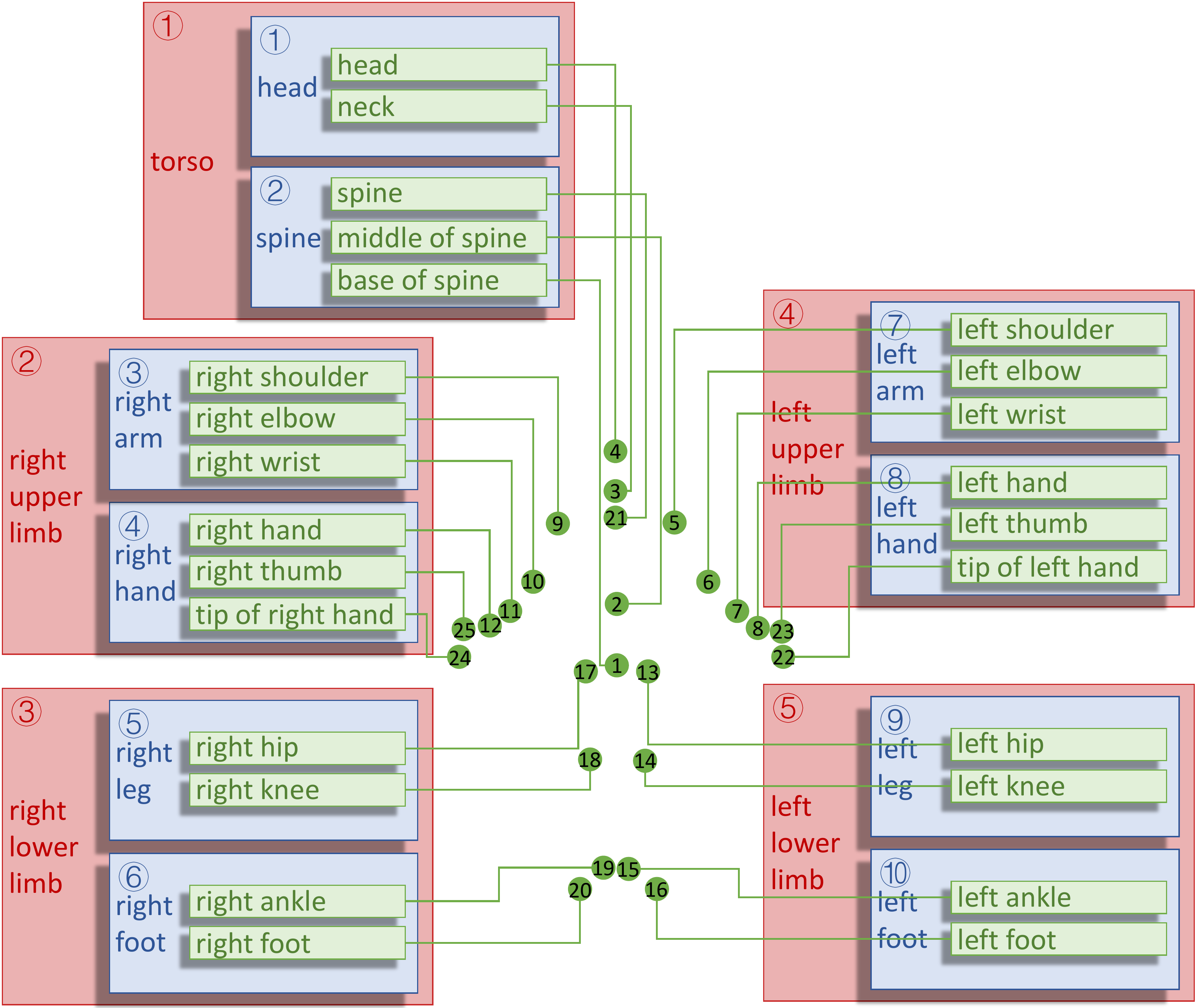}
    \caption{The illustration of body level, part level, and joint level. The red box is body level, the blue box is part level, and the green box is joint level.}
    \label{levels}
\end{figure}

Skeleton Pyramid transforms the input features $f_{j}$ and $f_m$ to three-level (i.e., body level, part level, and joint level) features involved in human body structure for learning coarse-to-fine semantic information. Based on the biological structure of the human body, $N$ joint-level nodes are spatially clustered into $P$ part-level nodes (e.g., hand, arm, foot), each of which includes several joint-level nodes. And then, $P$ part-level nodes are further clustered into $B$ body-level nodes (e.g., left upper limb, left lower limb, torso), each of which contains two or more part-level nodes, as shown in Figure~\ref{levels}. By passing through the Skeleton Pyramid, the joint features $f_{j}\in \mathbb{R}^{M\times C\times T\times N}$ are directly transformed to the joint-level joint features $g_j\in\mathbb{R}^{M\times C\times T\times N}$ without any operation; the joint-level joint features $g_j\in\mathbb{R}^{M\times C\times T\times N}$ are transformed to the part-level joint features $h_j\in\mathbb{R}^{M\times C\times T\times P}$ by squeezing $N$-dimension to $P$-dimension with the averaging operation; and the the part-level joint features $h_j\in\mathbb{R}^{M\times C\times T\times P}$ are further transformed to the body-level joint features $z_j\in\mathbb{R}^{M\times C\times T\times B}$ by squeezing $P$-dimension to $B$-dimension with the averaging operation. The above process can be formulated by a pyramid generator $\Lambda(\cdot)$, as follows, 
\begin{equation}
    <g_j, h_j, z_j> =  \Lambda(f_{j})
\end{equation}
In the same way, the motion features $f_m$ can be also transformed to the joint-level motion features $g_m$, part-level motion features $h_m$, and body-level motion features $z_m$ via the pyramid generator $\Lambda(\cdot)$.

\subsection{Pyramid Polymerizing Attention (PPA)}

To complement coarse-to-fine semantic information, Pyramid Polymerizing Attention (PPA) is designed to generate body-level, part-level, and joint-level polymerizing features by polymerizing body-level, part-level, and joint-level attention maps from body-to-part-to-joint level.
As shown in Figure~\ref{framework}, body-level features $z_{j/m}$ are first converted to the corresponding query features $Q$, key features $K$, and value features $V$ via multiple linear projection layers:
\begin{equation}
    Q, K, V = W^Qz_{j/m}, W^Kz_{j/m}, W^Vz_{j/m}
\end{equation}
where $W^Q$, $W^K$, and $W^V \in \mathbb{R}^{STC_e\times TC}$; $Q$, $K$, and $V \in \mathbb{R}^{M\times S\times B\times TC_e}$; $S$ denotes the parameter of multi-head, $C_e$ is a smaller number of channels than $C$ to reduce computation ($C=S\times C_e$). 
We utilize a body-level attention map $\mathcal{A}_z$ to assign weights to various body-level nodes and emphasize the crucial body-level information in a self-attention way, as follows,
\begin{equation}
    \mathcal{A}_z = \text{tanh}(\frac{QK^\top}{\sqrt{TC_e}})
\end{equation}
where $\mathcal{A}_z \in \mathbb{R}^{M\times S\times B\times B}$, $\top$ denotes the transpose operator. In contrast to the original self-attention mechanism, we choose ${\text{tanh}}(\cdot)$ instead of ${\text{softmax}}(\cdot)$ to make each element value in the generated attention map not only be positive, following~\cite{shi2020decoupled}. Meanwhile, the part-level attention map $\mathcal{A}_h \in \mathbb{R}^{M\times S\times P\times P}$ and joint-level attention map $\mathcal{A}_g \in \mathbb{R}^{M\times S\times N\times N}$ are obtained in the similar way. Here, $\mathcal{A}_z$, $\mathcal{A}_h$ and $\mathcal{A}_g$ focus on key information at the body level, part level, and joint level, respectively.

\begin{figure}
    \centering
    \includegraphics[width=3.4in]{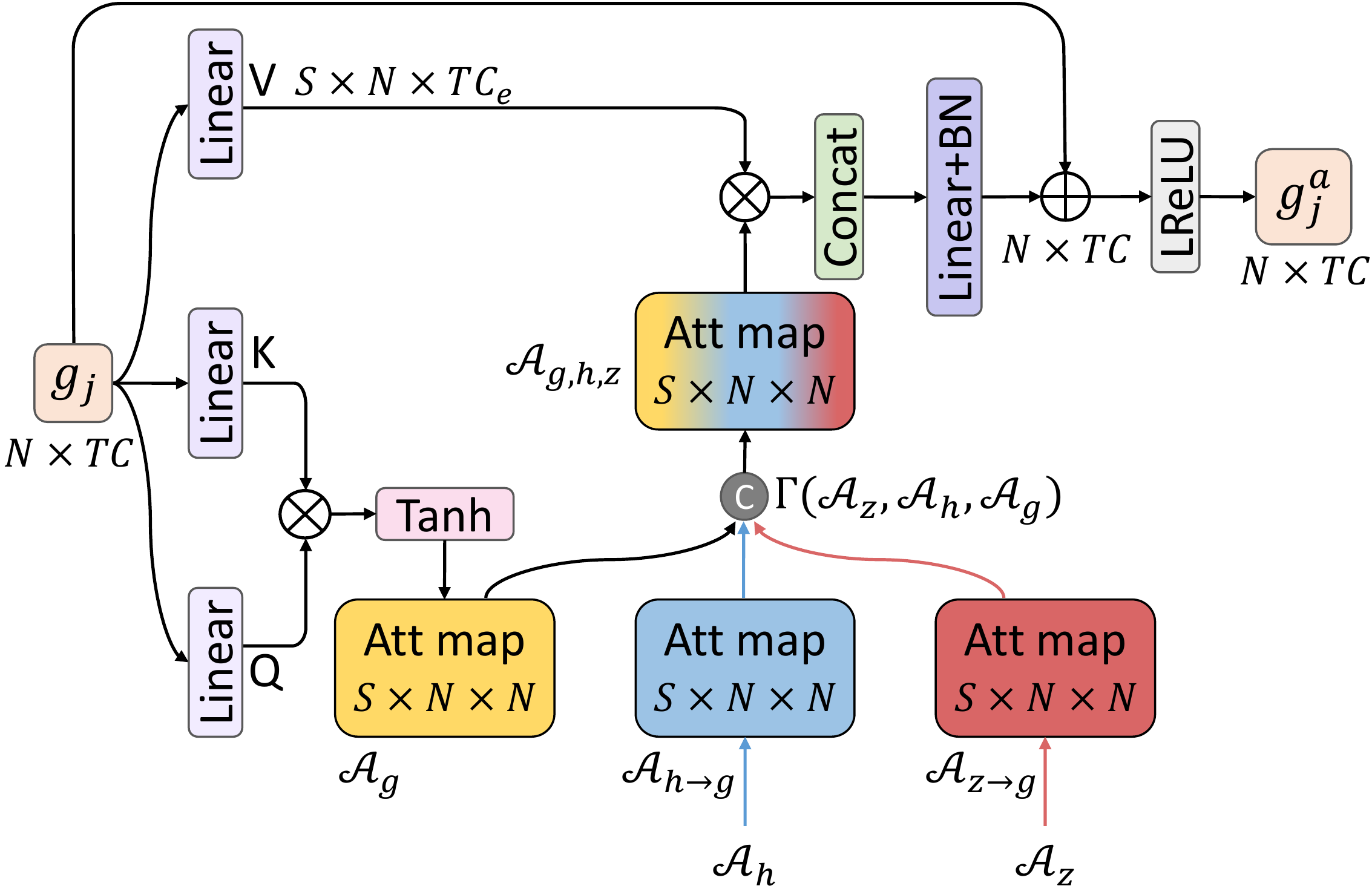}
    \caption{The process of producing the joint-level polymerizing features $g_j^a$. $g_j$ is joint-level features in joint modality. Linear is the linear projection layer. $\alpha$ and $\beta$ are hyperparameters. $\otimes$ denotes the matrix multiplication. BN is the batch norm. $\oplus$ denotes the residual connection addition. LReLU is the leaky ReLU activation function.}
    \label{attention}
\end{figure}

First, we can directly calculate the body-level polymerizing features $z_{j/m}^a$ by the following equation,
\begin{equation}
    z_{j/m}^a = \sigma(\varPsi(\text{concat}(\mathcal{A}_zV)) + z_{j/m})
\end{equation}
where $\varPsi(\cdot)$ is a Feed Forward Network (FFN) consisting of a linear layer with a batch norm layer, $\sigma(\cdot)$ is the activation function (e.g., leaky ReLU). Second, we calculate the part-level polymerizing features $h_{j/m}^a$ that polymerizes from body-to-part level semantic information by a polymerization generator $\Upsilon(\cdot)$, as follows, 
\begin{equation}
    h_{j/m}^a = \sigma(\varPsi(\text{concat}(\Upsilon(\mathcal{A}_{z},\mathcal{A}_{h})V)) + h_{j/m})
    \label{eq5}
\end{equation}
In Eq.~\eqref{eq5}, the polymerization generator $\Upsilon(\mathcal{A}_{z},\mathcal{A}_{h})$ aims to polymerize the body- and part-level attention maps for obtaining the part-body polymerizing attention map $\mathcal{A}_{h,z}$. Its implementation process is detailed as follows,
\begin{equation}
\begin{aligned}
\label{fill_1}
    &\mathcal{A}_{z} =\{\omega_{m,n}\}, 1\le m\le B, 1\le n\le B \\
    &\upsilon_{i,j} = \omega_{\phi(i),\phi(j)}, 1\le \phi(i)\le B, 1\le \phi(j)\le B \\
    &\mathcal{A}_{z\rightarrow h} =\{\upsilon_{i,j}\}, 1\le i\le P, 1\le j\le P \\
    &\Upsilon(\mathcal{A}_{z},\mathcal{A}_{h}) = \mathcal{A}_h + \lambda\mathcal{A}_{z\rightarrow h}
\end{aligned}
\end{equation}
Here $\mathcal{A}_{z\rightarrow h} \in \mathbb{R}^{M\times S\times P\times P}$, $\lambda$ is a hyper-parameter, and $\phi(.)$ denotes the affiliation from the part-level node to the body-level node, that is, the index of the body-level node to which the part-level node belongs. Similarly, we calculate the joint-level polymerizing features $g_{j/m}^a$ that polymerizes from part-to-joint and body-to-joint level semantic information by a polymerization generator $\Gamma(\cdot)$, as follows, 
\begin{equation}
    g_{j/m}^a = \sigma(\varPsi(\text{concat}(\Gamma(\mathcal{A}_{z},\mathcal{A}_{h},\mathcal{A}_{g})V)) + g_{j/m})
    \label{eq10}
\end{equation}
In Eq.~\eqref{eq10}, the polymerization generator $\Gamma(\mathcal{A}_{z},\mathcal{A}_{h}, \mathcal{A}_{g})$ aims to polymerize the body-, part- and joint-level attention maps for obtaining the joint-part-body polymerizing attention map $\mathcal{A}_{g,h,z}$. Its implementation process is detailed as follows,

\begin{equation}
    \Gamma(\mathcal{A}_z,\mathcal{A}_h,\mathcal{A}_g) = \mathcal{A}_g + \alpha\mathcal{A}_{h\rightarrow g} + \beta\mathcal{A}_{z\rightarrow g}
\end{equation}
where $\alpha$ and $\beta$ are hyperparameters, the calculation procedures of $\mathcal{A}_{h\rightarrow g}$ and $\mathcal{A}_{z\rightarrow g}$ are similar to that of $\mathcal{A}_{z\rightarrow h}$, except that $\phi(.)$ denotes the affiliation from the joint-level node to the part-level node or the body-level node. The whole procedure is shown in Figure~\ref{attention}.

\subsection{Coarse-to-fine Contrastive Loss (CCL)}
To facilitate the Encoder to learn the body-level, part-level, and joint-level complementary semantic information in joint and motion modalities from a large amount of unlabeled data, we design the new Coarse-to-fine Contrastive Loss (CCL). In CCL, body-level contrast loss, part-level contrast loss, and joint-level contrast loss are used to jointly measure the similarity between body-level, part-level, and joint-level contrasting features of joints and motions.

Specifically, as shown in Figure~\ref{framework}, body-level polymerizing features $z_{j/m}^a$ , part-level polymerizing features $h_{j/m}^a$ and joint-level polymerizing features $g_{j/m}^a$ are first converted to body-level contrasting features $z_{j/m}^p$ , part-level contrasting features $h_{j/m}^p$ and joint-level contrasting features $g_{j/m}^p$ in the contrastive space by Pooling and MLP for boosting the obtained representations. Then, body-level contrast loss $\mathcal{L}^z$ of $z_j^p$ and $z_m^p$ is computed as follows,
\begin{equation}
\label{eq_loss}
    \mathcal{L}^z = -\frac{1}{2M}\sum_{i=1}^{2M}\log\frac{\exp(\text{sim}(u_i, u_{\hat{i}})/\tau)}{\sum\nolimits_{k=1}^{2M}\mathds{1}_{[k\not=i]}\exp(\text{sim}(u_i,u_k)/\tau)}
\end{equation}
where $u_i$, $u_{\hat{i}} \in \mathcal{U}$, $\mathcal{U}=z_j^p \cup z_m^p$ ($\mathcal{U} \in \mathbb{R}^{2M\times C'\times 1\times 1}$), $u_i$ and $u_{\hat{i}}$ denote the representations from the same data yet different modalities, namely one from joint modality while another from motion modality; $\text{sim}(u_i,u_{\tilde{i}})=u_i^\top u_{\tilde{i}}/\| u_i\| \| u_{\tilde{i}}\|$; $\mathds{1}\in \{0,1\}$ is used to indicate whether $k$ and $i$ are equal; and $\tau$ is a hyperparameter. Similar to Eq.~(\ref{eq_loss}), the part-level contrast loss $\mathcal{L}^h$ of $h_j^p$ and $h_m^p$, and joint-level contrast loss $\mathcal{L}^g$ of $g_j^p$ and $g_m^p$ are also formulated as follows,
\begin{equation}
\label{eq_loss_h}
    \mathcal{L}^h = -\frac{1}{2M}\sum_{i=1}^{2M}\log\frac{\exp(\text{sim}(o_i, o_{\hat{i}})/\tau)}{\sum\nolimits_{k=1}^{2M}\mathds{1}_{[k\not=i]}\exp(\text{sim}(o_i,o_k)/\tau)}
\end{equation}
\begin{equation}
\label{eq_loss_g}
    \mathcal{L}^g = -\frac{1}{2M}\sum_{i=1}^{2M}\log\frac{\exp(\text{sim}(r_i, r_{\hat{i}})/\tau)}{\sum\nolimits_{k=1}^{2M}\mathds{1}_{[k\not=i]}\exp(\text{sim}(r_i,r_k)/\tau)}
\end{equation}
where $o_i$, $o_{\hat{i}} \in \mathcal{O}$, $\mathcal{O}=h_j^p \cup h_m^p$; $r_i$, $r_{\hat{i}} \in \mathcal{R}$, $\mathcal{R}=g_j^p \cup g_m^p$ ($\mathcal{O}$ and $\mathcal{R} \in \mathbb{R}^{2M\times C'\times 1\times 1}$).
Finally, the designed Coarse-to-fine Contrastive Loss (CCL) (denoted by $\mathcal{L}_{con}$) can be defined as follows,
\begin{equation}
    \mathcal{L}_{con} = \mathcal{L}^z + \mathcal{L}^h + \mathcal{L}^g
\end{equation}

\subsection{Training Objective}
For the skeleton-based action recognition task in the semi-supervised scenario, we train the whole model jointly by contrastive loss on unlabeled data and the recognition loss on labeled data. For such labeled data, the joint features and motion features are obtained through the Encoder, and then input into the Average Pooling (AP), a Fully Connected layer (FC), and softmax in turn to obtain the classification prediction label $\hat{y}$, which is trained by the recognition loss (e.g., cross-entropy loss), as follows,
\begin{equation}
    \hat{y} = \text{softmax}(\text{FC}(\text{AP}(f_j))+\text{FC}(\text{AP}(f_m)))
\end{equation}
\begin{equation}
    \mathcal{L}_{reg} = -y^{T}\log(\hat{y})
\end{equation}
where $y$ is the ground-truth label. Overall, the training objective of the proposed PSP Learning is $\mathcal{L}$, as follows,
\begin{equation}
    \mathcal{L} = \mathcal{L}_{con} + \mathcal{L}_{reg}
\end{equation}
Algorithm~1 summarizes the main implementations of PSP Learning.

 \begin{figure}[!t]
		\label{alg:PSPL}
		\renewcommand{\algorithmicrequire}{\textbf{Input:}}
        \removelatexerror
		\begin{algorithm}[H]
			\caption{Pyramid Self-attention Polymerization Learning (PSP Learning)}
			\begin{algorithmic}
			    \REQUIRE 
			    \STATE $\mathcal{X}_j$, $\mathcal{X}_m$ : joint data, and motiotn data
			    \STATE $K$ : total optimization steps
			    \STATE $\tau$, $\lambda$, $\alpha$, $\beta$ : hyperparameter
			    \STATE $y$ : the ground-truth label
		        \FOR{$k=1$ \TO $K$}
		        \STATE $f_j$, $f_m$ = Encoder($\mathcal{X}_j$), Encoder($\mathcal{X}_m$)
		        \STATE
		        \STATE $//$ {\textit {Skeleton Pyramid}}
		        \STATE $<z_{j/m}, h_{j/m}, g_{j/m}> = \Lambda(f_{{j/m}})$
                \STATE
                \STATE $//$ {\textit {Pyramid Polymerizing Attention (PPA)}}
                \begin{equation}
                    \begin{aligned}
                        &z_{j/m}^a = \sigma(\varPsi(\text{concat}(\mathcal{A}_zV)) + z_{j/m}) \notag \\ 
                        &h_{j/m}^a = \sigma(\varPsi(\text{concat}(\Upsilon(\mathcal{A}_{z},\mathcal{A}_{h})V)) + h_{j/m}) \notag \\ 
                        &g_{j/m}^a = \sigma(\varPsi(\text{concat}(\Gamma(\mathcal{A}_z,\mathcal{A}_h,\mathcal{A}_g)V)) + g_{j/m}) \notag
                    \end{aligned}
                \end{equation}
                \STATE
                \STATE $//$ {\textit {Coarse-to-fine Contrastive Loss (CCL)}}
                \begin{equation}
                    \begin{aligned}
                        &z_{j/m}^p = \text{MLP}(\text{Pooling}(z_{j/m}^a)) \notag \\
                        &h_{j/m}^p = \text{MLP}(\text{Pooling}(h_{j/m}^a)) \notag \\
                        &g_{j/m}^p = \text{MLP}(\text{Pooling}(g_{j/m}^a)) \notag \\ \\
                        &\mathcal{L}^z = -\frac{1}{2M}\sum_{i=1}^{2M}\log\frac{\exp(\text{sim}(u_i, u_{\hat{i}})/\tau)}{\sum\nolimits_{k=1}^{2M}\mathds{1}_{[k\not=i]}\exp(\text{sim}(u_i,u_k)/\tau)} \notag \\ \\
                        &\mathcal{L}^h = -\frac{1}{2M}\sum_{i=1}^{2M}\log\frac{\exp(\text{sim}(o_i, o_{\hat{i}})/\tau)}{\sum\nolimits_{k=1}^{2M}\mathds{1}_{[k\not=i]}\exp(\text{sim}(o_i,o_k)/\tau)} \notag \\ \\
                        &\mathcal{L}^g = -\frac{1}{2M}\sum_{i=1}^{2M}\log\frac{\exp(\text{sim}(r_i, r_{\hat{i}})/\tau)}{\sum\nolimits_{k=1}^{2M}\mathds{1}_{[k\not=i]}\exp(\text{sim}(r_i,r_k)/\tau)} \notag \\ \\
                        &\mathcal{L}_{con} = \mathcal{L}^z + \mathcal{L}^h + \mathcal{L}^g \notag
                    \end{aligned}
                \end{equation}
                \STATE
                \STATE $//$ {\textit {Training Objective}}
                \begin{equation}
                    \begin{aligned}
                        &\hat{y} = \text{softmax}(\text{FC}(\text{AP}(f_j))+\text{FC}(\text{AP}(f_m))) \notag \\
                        &\mathcal{L}_{reg} = -y^{T}\log(\hat{y}) \notag \\
                        &\mathcal{L} = \mathcal{L}_{con} + \mathcal{L}_{reg} \notag \\
                    \end{aligned}
                \end{equation}
                \STATE Update all parameters using Stochastic Gradient Descent (SGD) to minimize $\mathcal{L}$
		        \ENDFOR
			\end{algorithmic}
		\end{algorithm}
	\end{figure}

\section{Experiments}
\label{experiments}
\subsection{Dataset}
In order to comprehensively validate the performance of the proposed PSP Learning for the skeleton-based action recognition task in the semi-supervised scenario, extensive experiments are performed on two publicly available datasets, namely NTU RGB+D~\cite{shahroudy2016ntu} and Northwestern-UCLA~\cite{wang2014cross}.

\textbf{NTU RGB+D dataset~\cite{shahroudy2016ntu}.} It contains 56,578 skeleton sequences covering 60 common human action classes, collected from 40 different people and three Microsoft Kinetic v2 camera views. Each sequence includes multiple frames, each of which has no more than 2 bodies, and each body consists of 3D coordinates of 25 skeleton joints. For evaluating the performance of models under a unified standard, we also adopt two public benchmarks including Cross-Subject (CS) and Cross-View (CV) from~\cite{shahroudy2016ntu}.
In the Cross-Subject benchmark, 40,091 skeleton sequences from 20 people belong to the training set, while the skeleton sequences from all remaining people belong to the testing set. In the Cross-View benchmark, the training set (including 37,646 skeleton sequences) and testing set (including 18,932 skeleton sequences) are collected from camera views 2-3 and camera view 1, respectively. The semi-supervised configuration on this dataset is that training on 5\%, 10\%, 20\%, and 40\% labeled data and the rest unlabeled data~\cite{si2020adversarial}.

\textbf{Northwestern-UCLA (NW-UCLA) dataset~\cite{wang2014cross}.} It contains 1,494 skeleton sequences in 10 classes, collected from 10 people and three Microsoft Kinetic v1 camera views. In one skeleton sequence, each human body is represented by the coordinate information of 20 joints. Following the commonly-used evaluation benchmark~\cite{wang2014cross}, 1,018 skeleton sequences coming from camera views 1 and 2 belong to the training set, while 476 skeleton sequences coming from camera view 3 belong to the testing set. To be consistent with most previous works~\cite{si2020adversarial}, the semi-supervised configuration is that training on 5\%, 15\%, 30\%, and 40\% labeled data and the other unlabeled data.

\subsection{Setting and Implementation}
On the NTU RGB+D and NW-UCLA datasets, all skeleton sequences are sampled into 50 frames of fixed length along the temporal dimension. Following most common semi-supervised settings~\cite{si2020adversarial}, we use partial proportion labeled data to train the model, where the number of 5\% labeled skeleton sequences are approximately 1,980/1,860/50 on NTU RGB+D CS/NTU RGB+D CV/NW-UCLA, and the number of other proportions can be calculated accordingly.

In PSP Learning, the implementation of Encoder refers to~\cite{shi2020decoupled}, $B$ and $P$ are both set as 5 and 10 on the NTU RGB+D and NW-UCLA datasets. The number of self-attention heads in PPA is 4, namely $S=4$, referring to~\cite{shi2020decoupled}; $\lambda$, $\alpha$, and $\beta$ are set as 0.2, 0.12, and 0.24 on NTU RGB+D, as well as 0.2, 0.1, and 0.2 on NW-UCLA. $\tau$ in CCL is set as 0.07.

By combining the recognition loss and CCL, we train the semi-supervised learning-based PSP Learning on both labeled data and unlabeled data, and optimize all parameters via Stochastic Gradient Descent (SGD) with Nesterov momentum 0.9. On NTU RGB+D/NW-UCLA, the batch size, the learning rate, weight decay, warmup epoch~\cite{he2016deep}, and total training epochs are set as 64/128, 0.05/0.06, $0.5\times e^{-3}$/$0.1\times e^{-3}$, 5/20, and 120/300, respectively. The learning rate is divided by 10 in epoch 60 and 90 on NTU RGB+D, as well as in epoch 76 and 130 on NW-UCLA. We adopt the PyTorch framework to implement the proposed method and employ a Titan RTX GPU to run all experiments in the Linux environment. 

\begin{table*}[ht]
\renewcommand{\arraystretch}{1.3}
\caption{Recognition accuracies (\%) obtained by different methods on the NTU RGB+D dataset (Cross-Subject (CS) and Cross-View (CV)) with 5\%, 10\%, 20\%, and 40\% labeled data of training set. }
\label{table1}
\centering
\begin{tabular}{l||c|c||c|c||c|c||c|c}
\hline\hline
\multirow{2}{*}{Method}&
\multicolumn{2}{c||}{5\%} & \multicolumn{2}{c||}{10\%} & \multicolumn{2}{c||}{20\%} & \multicolumn{2}{c}{40\%} \\
\cline{2-9}
& CS & CV & CS & CV & CS & CV & CS & CV \\ \hline
S$^{4}$L \cite{zhai2019s4l} & 48.4  & 55.1  & 58.1  & 63.6 & 63.1  & 71.1  & 68.2  & 76.9 \\ \hline
Pseudolabels \cite{lee2013pseudo} & 50.9  & 56.3  & 58.4  & 65.8 & 63.9  & 71.2  & 69.5  & 77.7  \\ \hline
VAT \cite{miyato2018virtual} & 51.3  & 57.9  & 60.3  & 66.3 & 65.6  & 72.6  & 70.4  & 78.6 \\ \hline
VAT+EntMin \cite{grandvalet2005semi} & 51.7  & 58.3  & 61.4  & 67.5 & 65.9  & 73.3  & 70.8  & 78.9 \\ \hline
ASSL \cite{si2020adversarial} & 57.3  & 63.6  & 64.3  & 69.8 & 68.0  & 74.7  & 72.3  & 80.0 \\ \hline
AL+K \cite{li2020sparse} & 57.8 & - & 62.9 & - & - & - & - & - \\ \hline
X-CAR \cite{xu2022x} & 67.3 & 70.0 & 76.1 & 78.2 & 79.4 & 85.7 & 84.1 & 90.4 \\ \hline
AS-CAL \cite{rao2021augmented} & - & - & 52.2 & 57.3 & - & - & - & - \\ \hline
LongT GAN \cite{zheng2018unsupervised} & - & - & 62.0 & - & - & - & - & - \\ \hline
Holden et al. \cite{holden2015learning} & - & - & - & - & - & - & 72.9 & 81.1 \\ \hline
EnGAN-PoseRNN \cite{kundu2019unsupervised} & - & - & - & - & - & - & 78.7 & 86.5 \\ \hline
MS$^{2}$L \cite{lin2020ms2l} & - & - & 65.2 & - & - & - & - & - \\ \hline
Skeleton-Contrastive \cite{thoker2021skeleton} & 59.6 & 65.7 & 65.9 & 72.5 & 70.8 & 78.2 & - & - \\ \hline
GL-Transformer \cite{kim2022global} & 64.5 & 68.5 & 68.6 & 74.9 & - & - & - & - \\ \hline
CPM \cite{zhang2022contrastive} &	- & - & 73.0 & 77.1 & - & - & - & - \\ \hline
3s-Colorization \cite{yang2021skeleton} & 65.7 & 70.3 & 71.7 & 78.9 & 76.4 & 82.7 & 79.8 & 86.8 \\ \hline
3s-CrosSCLR \cite{li20213d} & - & - & 74.4 & 77.8 & - & - & - & - \\ \hline
CMD \cite{mao2022cmd} & 71.0 & 75.3 & 75.4 & 80.2 & 78.7 & 84.3 & - & - \\ \hline \hline
PSP Learning (Ours) & \textbf{72.2} & \textbf{76.3} & \textbf{78.1} & \textbf{82.1} & \textbf{82.6} & \textbf{86.9} & \textbf{85.4} & \textbf{90.8} \\ \hline
\end{tabular}
\end{table*}

\begin{table}[!t]
\renewcommand{\arraystretch}{1.3}
\caption{Recognition accuracies (\%) obtained by different methods on NW-UCLA with 5\%, 15\%, 30\%, and 40\% labeled data of training set.}
\label{table2}
\centering
\begin{tabular}{l||c|c|c|c}
  \hline\hline
  Method & 5\% & 15\% & 30\% & 40\% \\ \hline
  S$^{4}$L \cite{zhai2019s4l} & 35.3 & 46.6 & 54.5 & 60.6 \\ \hline
 Pseudolabels \cite{lee2013pseudo} & 35.6 & 48.9 & 60.6 & 65.7 \\ \hline
  VAT \cite{miyato2018virtual} & 44.8 & 63.8 & 73.7 & 73.9 \\ \hline
  VAT+EntMin \cite{grandvalet2005semi} & 46.8 & 66.2 & 75.4 & 75.6 \\ \hline
 ASSL \cite{si2020adversarial} & 52.6 & 74.8 & 78.0 & 78.4 \\ \hline
  AL+K \cite{li2020sparse} & 63.6 & 76.8 & 77.2 & 78.9 \\ \hline
  X-CAR \cite{xu2022x} & 68.7 & 77.5 & \textbf{80.9} & 83.1 \\ \hline
 GL-Transformer \cite{kim2022global} & 58.5 & - & - & - \\ \hline
  MS$^{2}$L \cite{lin2020ms2l} & - & 60.5 & - & - \\ \hline\hline
  PSP Learning (Ours) & \textbf{69.2} & \textbf{77.8} & {80.6} & \textbf{83.6} \\ \hline
\end{tabular}
\end{table}

\subsection{Experimental Result and Analysis}
The performance comparisons between the previously-related methods and the proposed PSP Learning for skeleton-based action recognition are illustrated in Table~\ref{table1} and Table~\ref{table2}. As a whole, the proposed PSP Learning performs well for the semi-supervised recognition task compared with the alternatives.

On the NTU RGB+D dataset, the comparative methods include semi-supervised learning-based methods~\cite{zhai2019s4l,lee2013pseudo,miyato2018virtual,grandvalet2005semi,si2020adversarial,li2020sparse,xu2022x}, and unsupervised learning-based methods~\cite{rao2021augmented,zheng2018unsupervised,holden2015learning,kundu2019unsupervised,lin2020ms2l,thoker2021skeleton,yang2021skeleton,li20213d,mao2022cmd,zhang2022contrastive,kim2022global}. In comparison to these semi-supervised methods, PSP Learning (with accuracy of 78.1\%) improves by 13.8\% over the state-of-the-art ASSL (with accuracy of 64.3\%) on CS with 10\% labeled data. Here, ASSL aims to capture more semantic information by exploring neighborhood consistency in a common modality. PSP Learning captures more semantic information by learning the consistency across joint and motion modalities. Compared with some representative unsupervised learning-based methods, the recognition accuracy of 82.6\% achieved by PSP Learning gains 3.9\% higher than that of the state-of-the-art CMD on CS with 20\% labeled data. Here, CMD mines semantic information in multiple views. PSP Learning considers mining rich semantic information in multiple modalities and multiple granularities.

On the NW-UCLA dataset, compared with the semi-supervised learning-based methods~\cite{zhai2019s4l,lee2013pseudo,miyato2018virtual,grandvalet2005semi,si2020adversarial,li2020sparse,xu2022x}, PSP Learning is comparable to the SOTA method (i.e., X-CAR), and outperforms the rest alternatives. Specifically, X-CAR explores the consistent action representations between joint data and the learnable augmentation data via contrastive learning. PSP Learning explores the consistent multi-granularity representations between the joint and motion data at joint, part, and body levels. Compared with unsupervised learning-based method~\cite{lin2020ms2l,kim2022global}, PSP Learning (with accuracy of 77.8\%) has 17.3\% improvement over the SOTA method (i.e., MS$^{2}$L~\cite{lin2020ms2l}) on the semi-supervised setting (with 15\% labeled data). Here, MS$^{2}$L learns action representations by the contrastive consistency between original data and augmented data. PSP Learning learns multiple-granularity action representations via the coarse-to-fine contrastive consistency at the joint, part, and body levels.

\begin{figure*}[!t]
    \centering
    \includegraphics[width=7.1in]{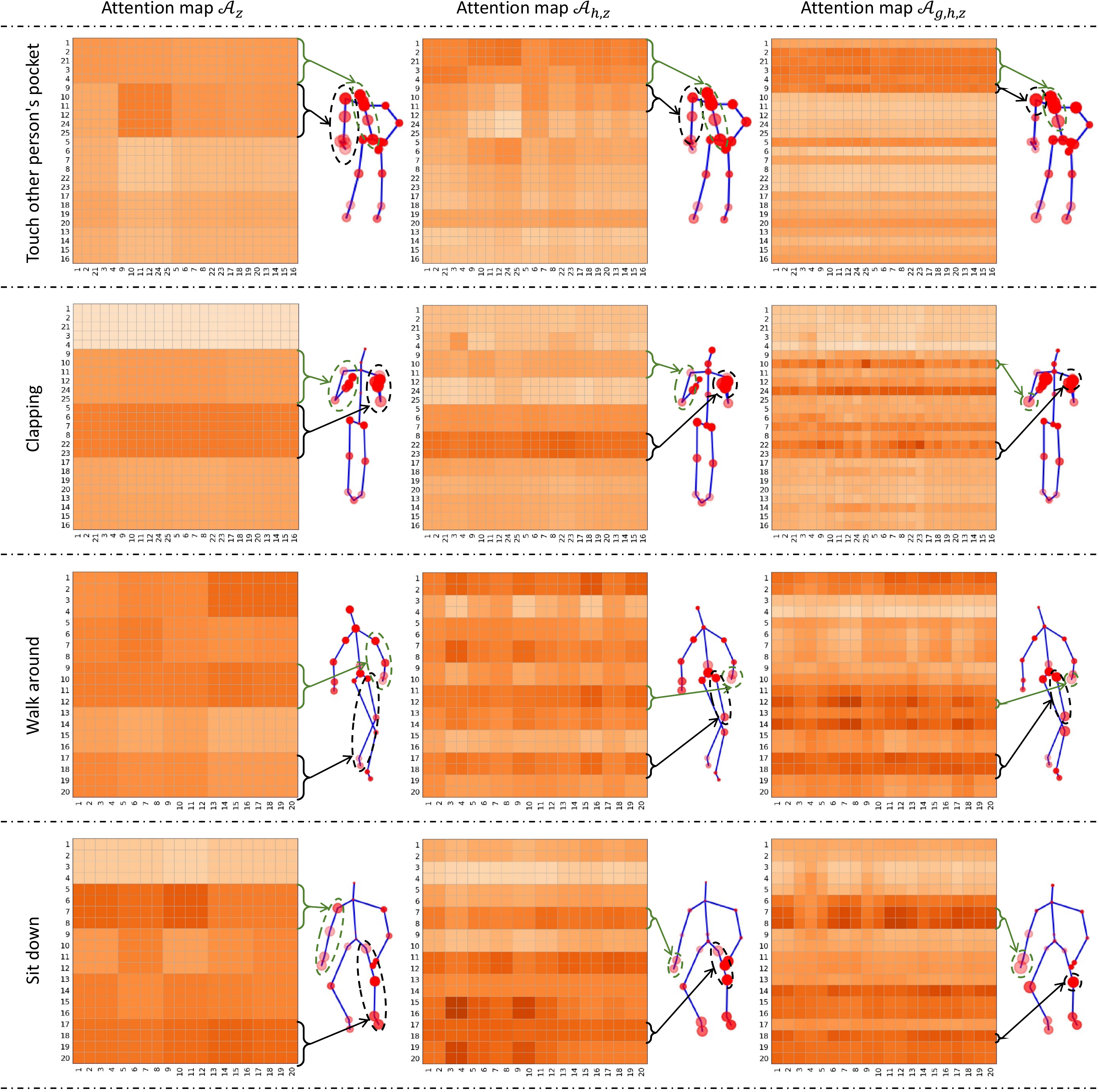}
    \caption{The visualization of attention maps $\mathcal{A}_z$, $\mathcal{A}_{h,z}$, and $\mathcal{A}_{g,h,z}$ in PPA for different action examples. The first two rows are from the NTU RGB+D dataset, and the last two rows are from the NW-UCLA dataset. Each row from left to right shows the attention maps $\mathcal{A}_z$, $\mathcal{A}_{h,z}$, and $\mathcal{A}_{g,h,z}$, respectively. Each attention map accompanies a skeleton visualization on the right. In the attention map, the darker color of the row, the more important the joint of the corresponding row is to the action. In the skeleton visualization, the larger size of the red circle, the more important the corresponding joint is to the action.}
    \label{fig:attention}
\end{figure*}

\subsection{Qualitative Analysis}
\subsubsection{Visualization of attention maps in PPA}
As the key module in PSP Learning, PPA finally outputs three types of attention maps $\mathcal{A}_z$, $\mathcal{A}_{h,z}$, and $\mathcal{A}_{g,h,z}$, which focus on the polymerizing of key semantic information from body level, body-to-part level, and body-to-part-to-joint level, respectively. In order to further understand the polymerizing process of coarse-to-fine information by PPA, we visualize attention maps $\mathcal{A}_z$, $\mathcal{A}_{h,z}$, and $\mathcal{A}_{g,h,z}$ of different examples on the NTU RGB+D and NW-UCLA datasets, as shown in Figure~\ref{fig:attention}.

Specifically, based on the polymerizing route $\mathcal{A}_z \rightarrow \mathcal{A}_{h,z} \rightarrow \mathcal{A}_{g,h,z}$, we can find that: i) The focus of actions ``Clapping", ``Walk around", and ``Sit down" shifts from the upper limb to the hand, then to the tip of the hand; ii) The focus of action ``Sit down" transforms from the lower limb to the leg, and then to the knee; iii) The focus of action ``Touch other person's pocket" progressively locates the position of the right shoulder and neck.
To sum up, the focus of key information is gradually refined with consistency from attention map $\mathcal{A}_z$ to $\mathcal{A}_{h,z}$, and further to $\mathcal{A}_{g,h,z}$.

\begin{figure*}[!t]
    \centering
    \subfloat[On PSP Learning w/o PPA w/o CCL.]{\includegraphics[scale=0.4]{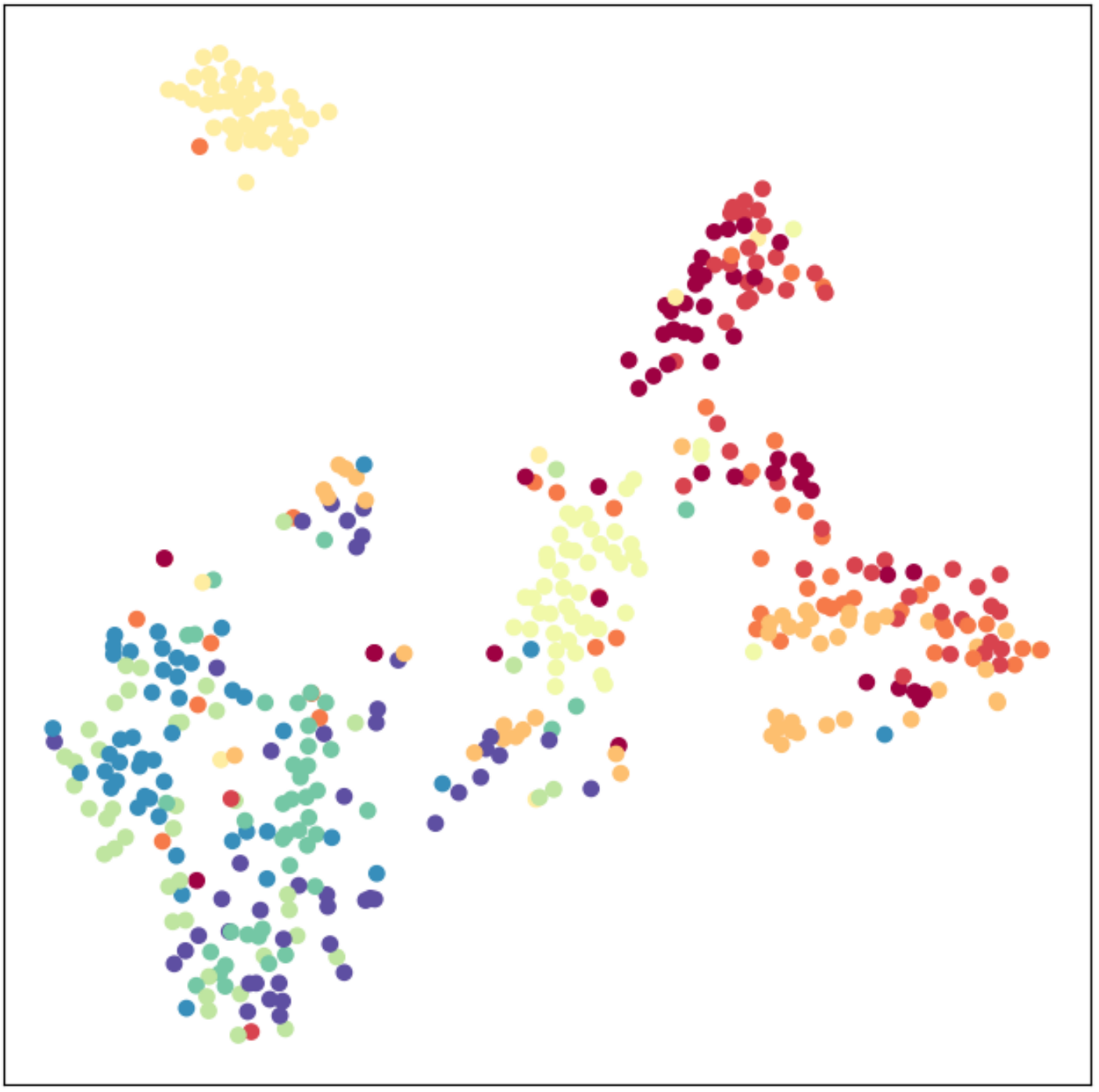}
    \label{fig_tsne_sup}}
    \hfil
    \subfloat[On PSP Learning w/o PPA w/ CCL.]{\includegraphics[scale=0.4]{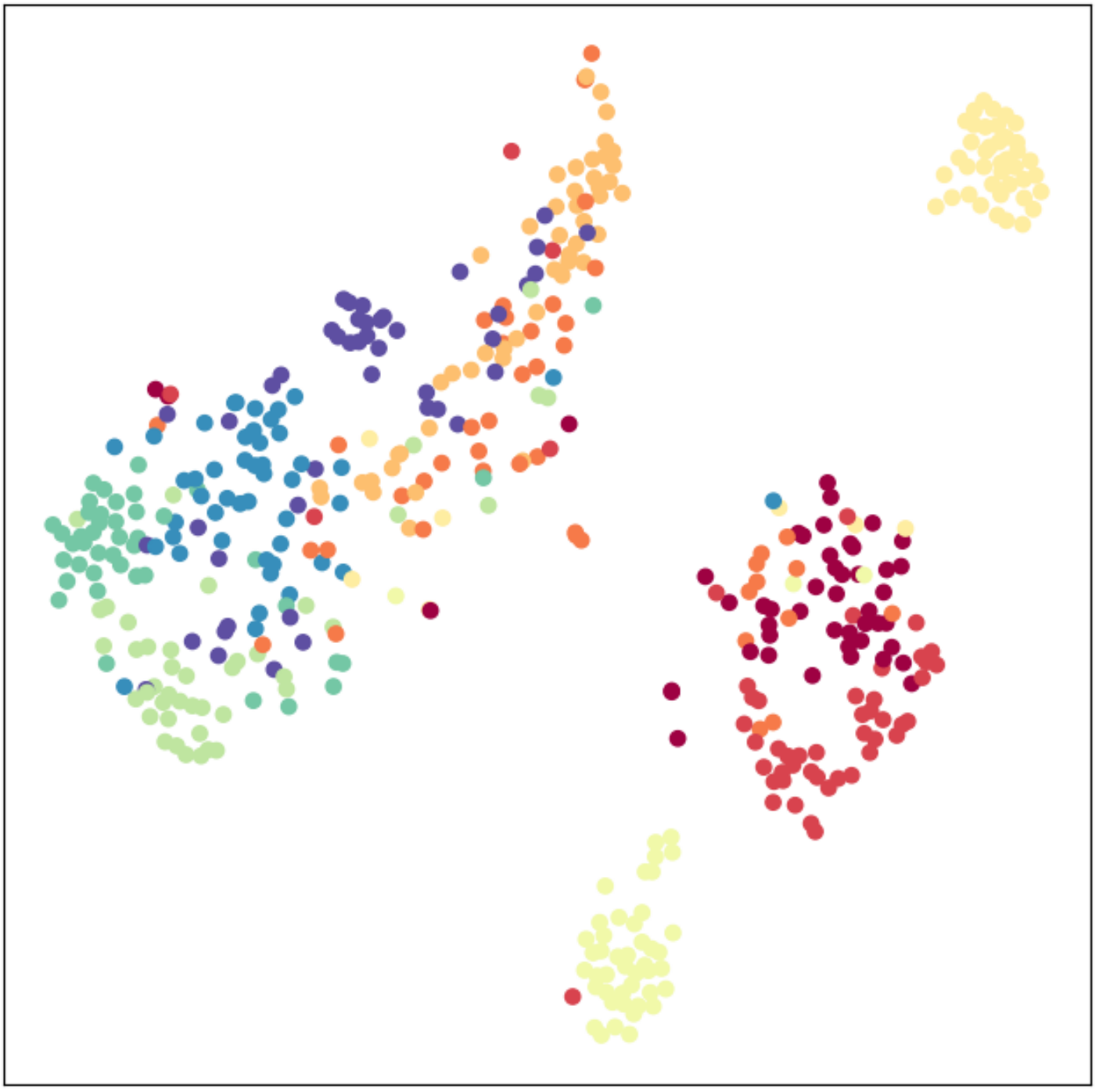}
    \label{fig_tsne_ccl}}
    \hfil
    \subfloat[On PSP Learning w/ PPA w/ CCL.]{\includegraphics[scale=0.4]{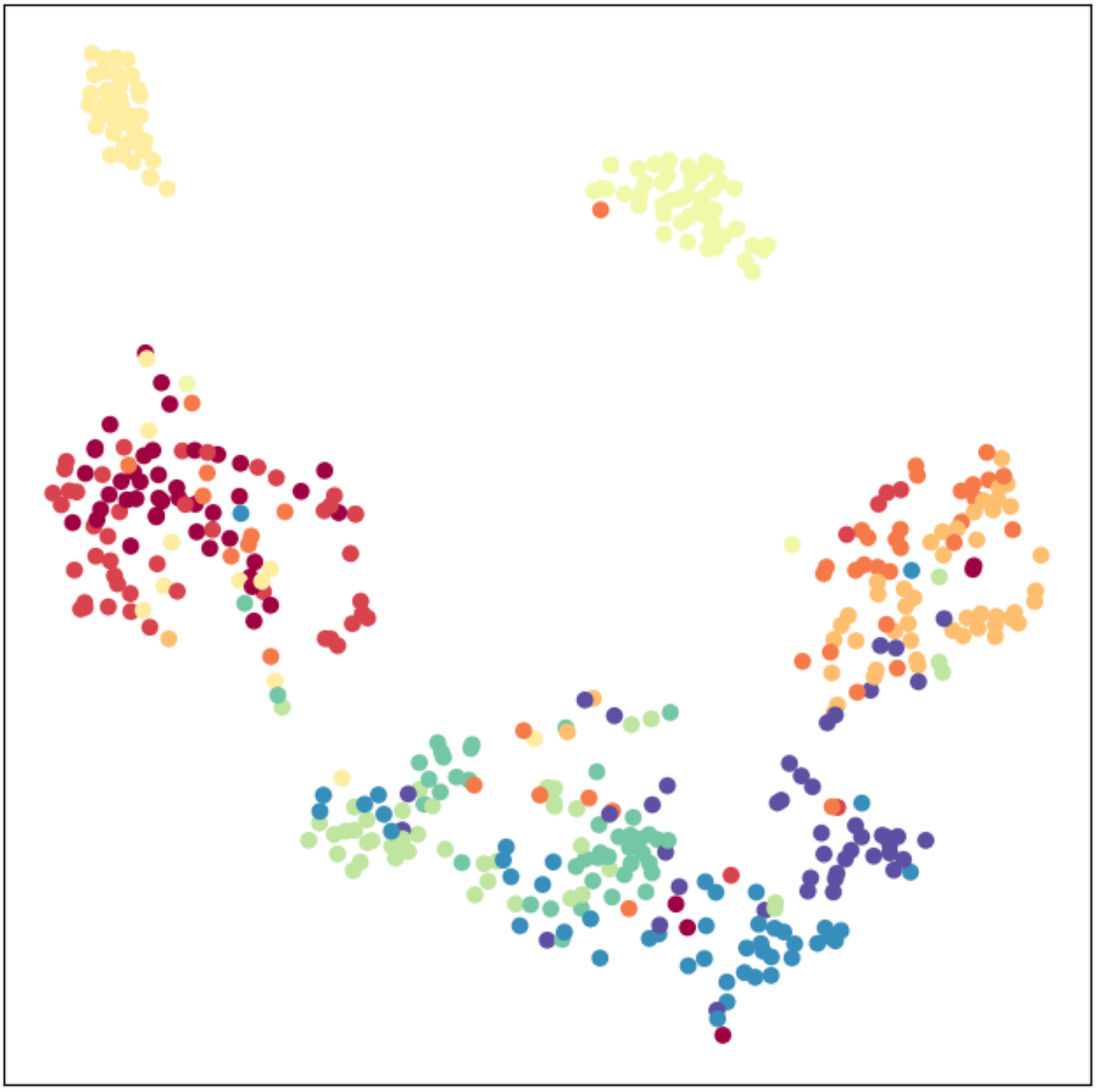}
    \label{fig_tsne_psp}}
    \caption{The t-SNE visualization of features obtained by different components on the NW-UCLA dataset. (a) t-SNE visualization of action features learned by PSP Learning without PPA and CCL; (b) t-SNE visualization of action features learned by PSP Learning without PPA; and (c) t-SNE visualization of action features learned by PSP Learning. Best view in the color PDF file.}
    \label{fig_tsne}
\end{figure*}

\subsubsection{Visualization of learned features}
In this work, to illustrate the ability of the representation learning of PSP Learning, we perform t-SNE visualizations to show the distribution difference of the action features learned by PSP Learning equipped with these different modules, i.e., Pyramid Polymerizing Attention (PPA), and Coarse-to-fine Contrastive Loss (CCL). At first, we set PSP Learning without PPA and CCL as one baseline, and show the visualized distribution of the learned features in Figure~\ref{fig_tsne_sup}. Compared with Figure~\ref{fig_tsne_sup}, the feature distribution learned in Figure~\ref{fig_tsne_ccl} is more aggregated in the same category, which verifies the effectiveness of CCL. Compared with Figure~\ref{fig_tsne_ccl}, the semantics of different classes of features learned in Figure~\ref{fig_tsne_psp} are more discriminative, which also proves the effectiveness of PPA.

\subsection{Ablation Studies}
Pyramid Polymerizing Attention (PPA) and Coarse-to-fine Contrastive Loss (CCL) are important modules in PSP Learning. Thus, we conduct ablation studies to verify the effectiveness of PPA and CCL in the semi-supervised recognition task.

\subsubsection{Effect of PPA}
\begin{figure}[!t]
    \centering
    \includegraphics[width=3.0in]{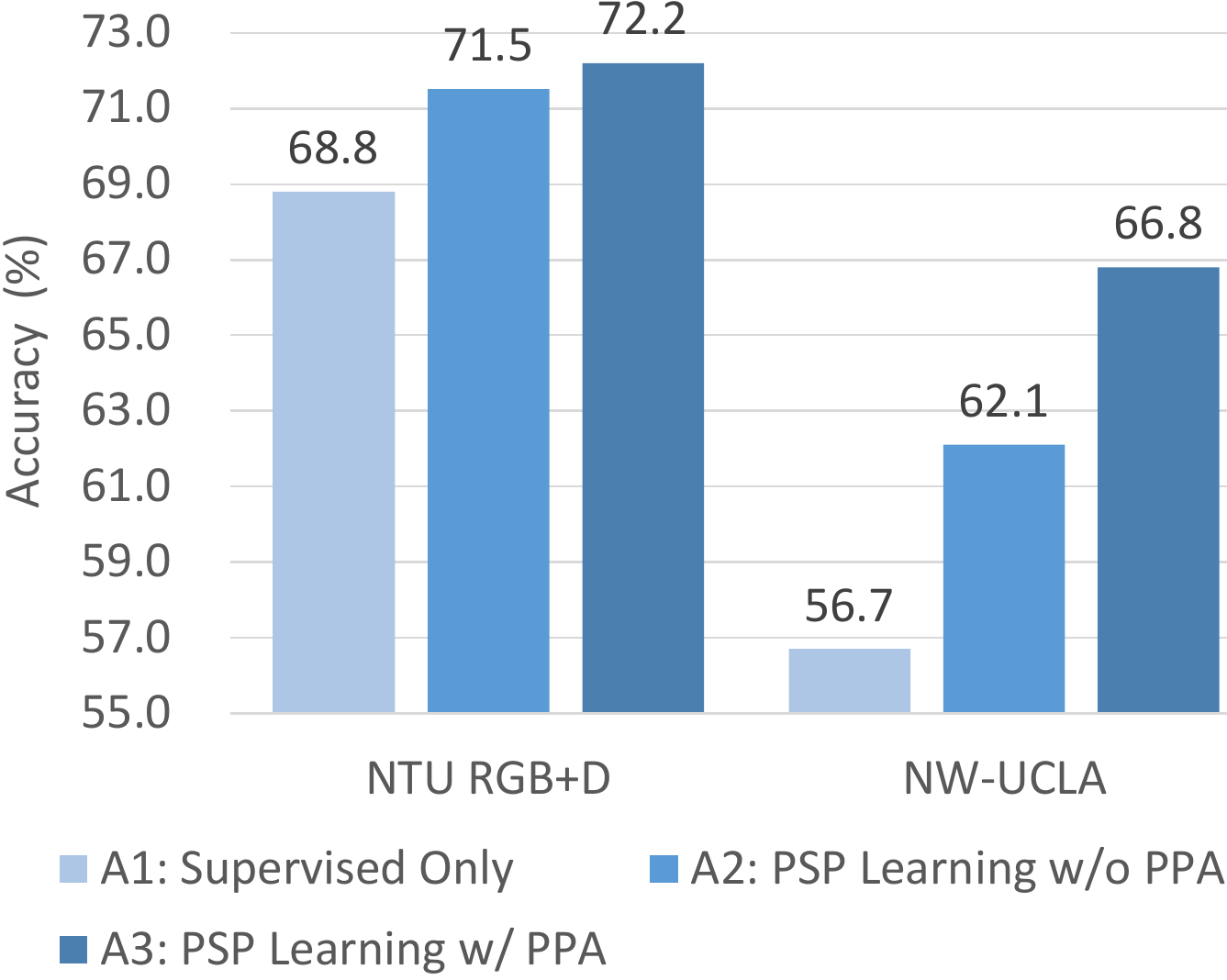}
    \caption{Accuracies (\%) obtained by different baselines (including Supervised Only, PSP Learning w/o PPA, and PSP Learning w/ PPA) on NTU RGB+D (CS) with 5\% labeled data and NW-UCLA with 5\% labeled data.}
    \label{PPA}
\end{figure}

As shown in Figure~\ref{PPA}, we compare three baselines (namely Supervised Only, PSP Learning w/o PPA, and PSP Learning w/ PPA) on NTU RGB+D (CS) with 5\% labeled data and NW-UCLA with 5\% labeled data. All baselines are defined as follows,
\begin{description}
\item[{\bf A1}] {\bf Supervised Only}. The whole framework only involves the Encoder and recognition classifier. Thus, the training of the Encoder with the recognition classifier only uses the labeled data. This aims to test the base performance by training the model with only labeled data. 
\item[{\bf A2}] {\bf PSP Learning w/o PPA}. By discarding PPA, action features output from the Skeleton Pyramid are directly input into CCL. This aims to test the superiority of PPA.
\item[{\bf A3}] {\bf PSP Learning w/ PPA}. This is the proposed PSP Learning.
\end{description}
We observe that A3 (PSP Learning w/ PPA) significantly improves by 10.1\% over A1 (i.e., Supervised Only) on NW-UCLA. Compared with A2 (i.e., PSP Learning w/o PPA), the performance improvements of A3 (i.e., PSP Learning w/ PPA) are 0.7\% and 4.7\% on NTU RGB+D and NW-UCLA, respectively. It is well demonstrated that PPA allied with contrastive learning is effective.

\subsubsection{Effect of CCL}
We evaluate the superiority of CCL by setting the following eight baselines in the ablation study.
\begin{description}
    \item[{\bf B1}] {\bf PSP Learning w/o CCL}. Its architecture is the same as that of Baseline A1. Since there is no CCL, its training is the supervised way on the labeled data. 
    \item[{\bf B2}] {\bf PSP Learning w/o part- and body-level contrast loss}. It only retains the joint-level branch, namely including joint-level attention, and joint-level contrast loss. The purpose is to test the base performance of joint-level contrastive learning.
    \item[{\bf B3}] {\bf PSP Learning w/o joint- and body-level contrast loss}. It only retains the part-level branch, namely including part-level attention, and part-level contrast loss. The purpose is to test the base performance of part-level contrastive learning.
    \item[{\bf B4}] {\bf PSP Learning w/o joint- and part-level contrast loss}. It only retains the body-level branch, namely including body-level attention, and body-level contrast loss. The purpose is to test the base performance of body-level contrastive learning.
    \item[{\bf B5}] {\bf PSP Learning w/o body-level contrast loss}. It discards the body-level branch, which can be seen as the merging of B2 and B3. The purpose is to test the importance of body-level contrast loss.
    \item[{\bf B6}] {\bf PSP Learning w/o part-level contrast loss}. It discards the part-level branch, which can be seen as the merging of B2 and B4. The purpose is to test the importance of part-level contrast loss.
    \item[{\bf B7}] {\bf PSP Learning w/o joint-level contrast loss}. It discards the joint-level branch, which can be seen as the merging of B3 and B4. The purpose is to test the importance of joint-level contrast loss.
    \item[{\bf B8}] {\bf PSP Learning (Ours)}. It is equal to the merging of B2, B3, and B4.
\end{description}
\begin{table}[!t]
\renewcommand{\arraystretch}{1.3}
\caption{Accuracies (\%) obtained by PSP Learning with different-level contrast losses on NTU RGB+D (CS) with 5\% labeled data.}
\label{ablation}
\centering
\begin{tabular}{c||c|c|c||c}
\hline\hline
\multirow{2}{*}{Baseline}&
\multicolumn{3}{c||}{CCL}&
\multirow{2}{*}{Accuracy}\\
\cline{2-4}
& Joint-level & Part-level & Body-level & \\ \hline
B1 & $\times$ & $\times$ & $\times$ & 68.8  \\ \hline
B2 & $\surd$ & $\times$ & $\times$ & 70.8  \\ \hline
B3 & $\times$ & $\surd$ & $\times$ & 70.6  \\ \hline
B4 & $\times$ & $\times$ & $\surd$ & 70.9  \\ \hline
B5 & $\surd$ & $\surd$ & $\times$ & 71.3  \\ \hline
B6 & $\surd$ & $\times$ & $\surd$ & 71.2  \\ \hline
B7 & $\times$ & $\surd$ & $\surd$ & 71.4  \\ \hline
B8 (Ours) & $\surd$ & $\surd$ & $\surd$ & {\bf 72.2} \\ \hline
\end{tabular}
\end{table}

Table~\ref{ablation} shows the recognition performance of different baselines. Compared with B1 (i.e., PSP Learning w/o CCL), there is a certain improvement of recognition performance for the other baselines (B2-B8) by equipping with various contrastive learning tricks. From the comparative results between B2-B4 and B5-B7, it can be concluded that integrating multiple (at least two) contrast losses can learn richer representations to benefit performance improvement. In short, B8 (the proposed PSP Learning) achieves 72.2\%, outperforming B1 (without any contrast loss), B4 (with only one contrast loss), and B7 (combining two contrast losses) by 3.4\%, 1.3\%, and 0.8\%, which demonstrates that the idea of incorporating multi-granularity contrastive losses is feasible.

\begin{figure*}[!t]
    \centering
    \subfloat[On PSP Learning w/o CCL.]{\includegraphics[scale=0.3]{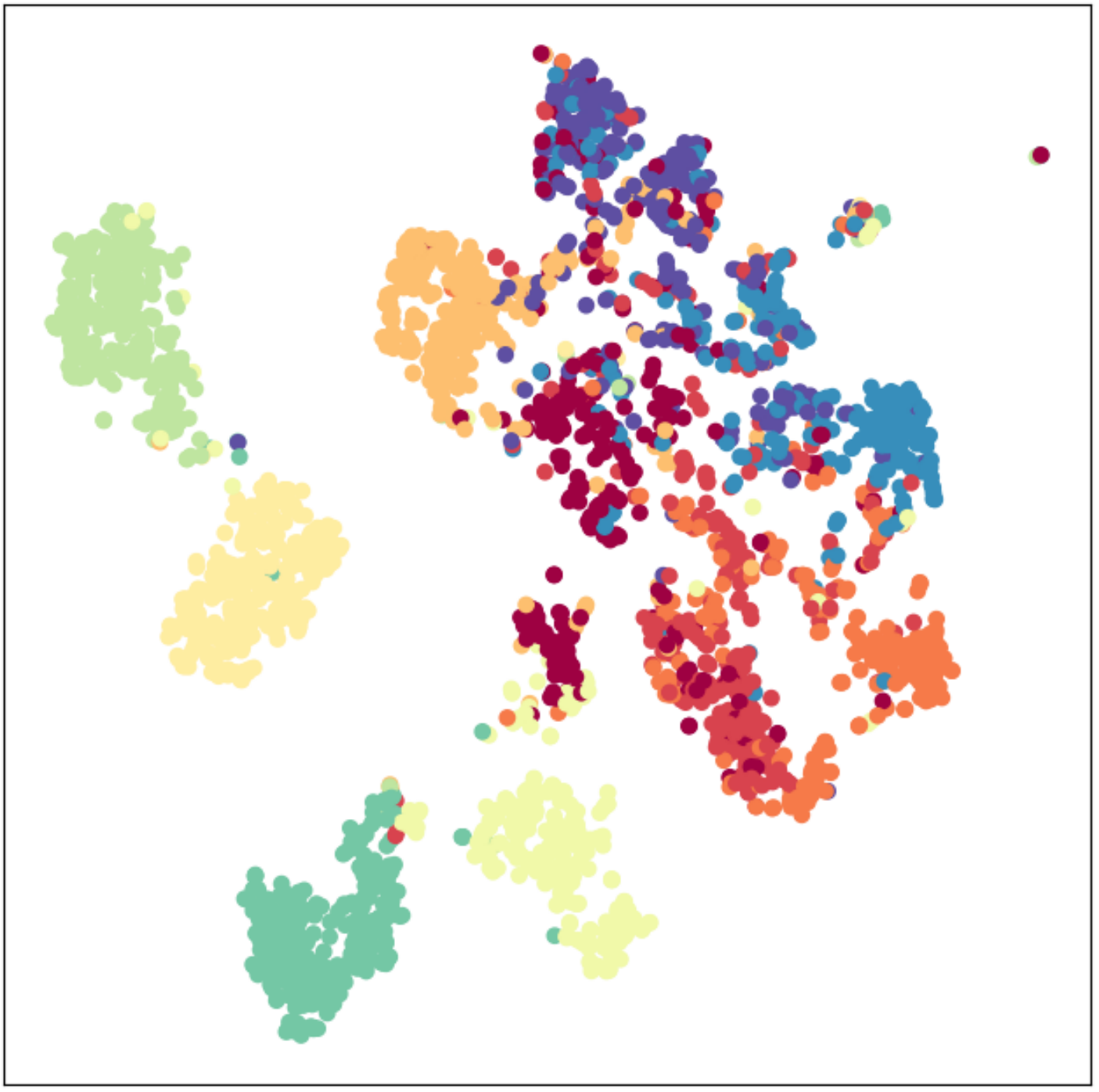}
    \label{fig_tsne_b1}}
    \hfil
    \subfloat[On PSP Learning w/o part- and body-level contrast loss.]{\includegraphics[scale=0.3]{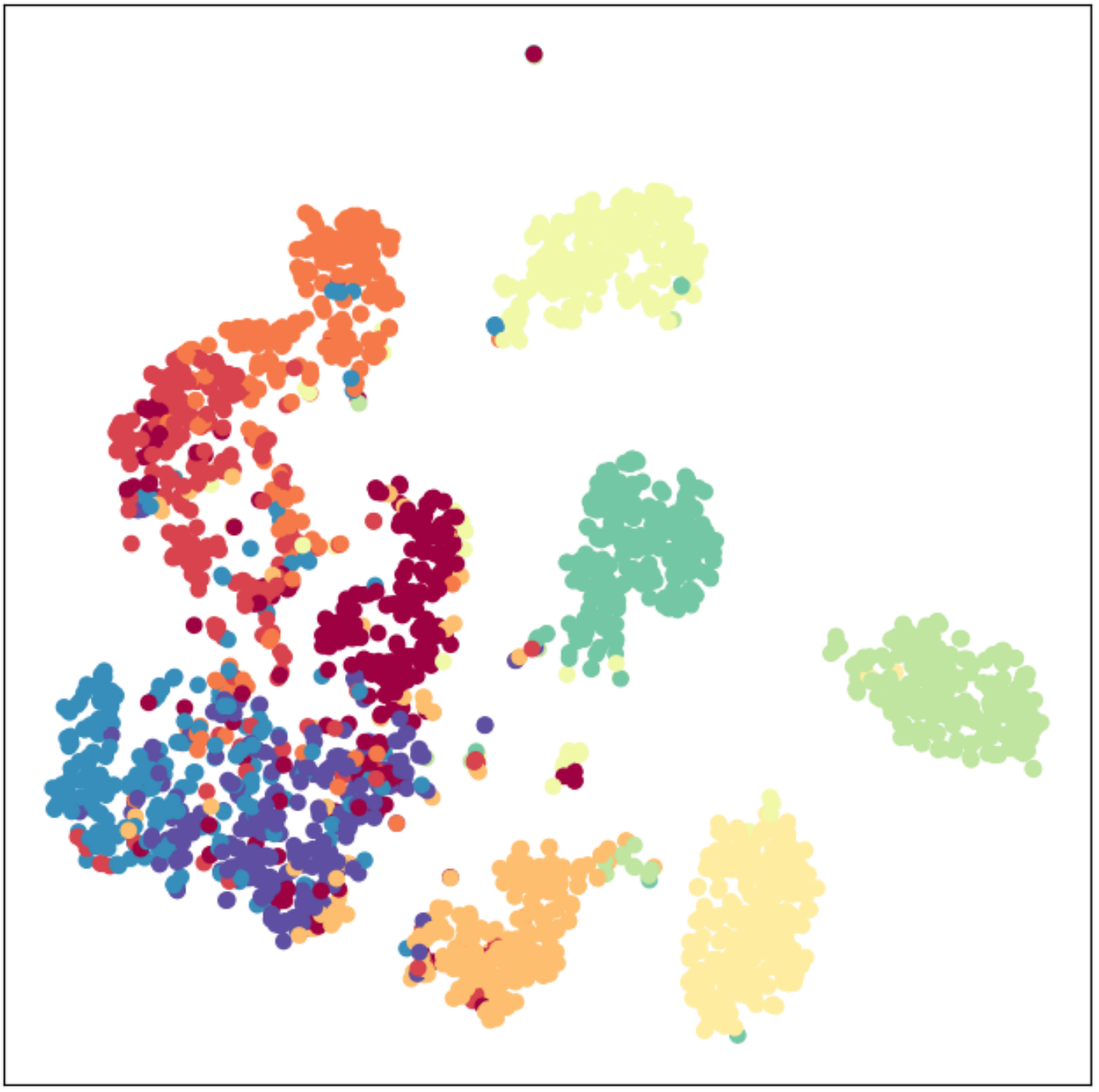}
    \label{fig_tsne_b2}}
    \hfil
    \subfloat[On PSP Learning w/o joint- and body-level contrast loss.]{\includegraphics[scale=0.3]{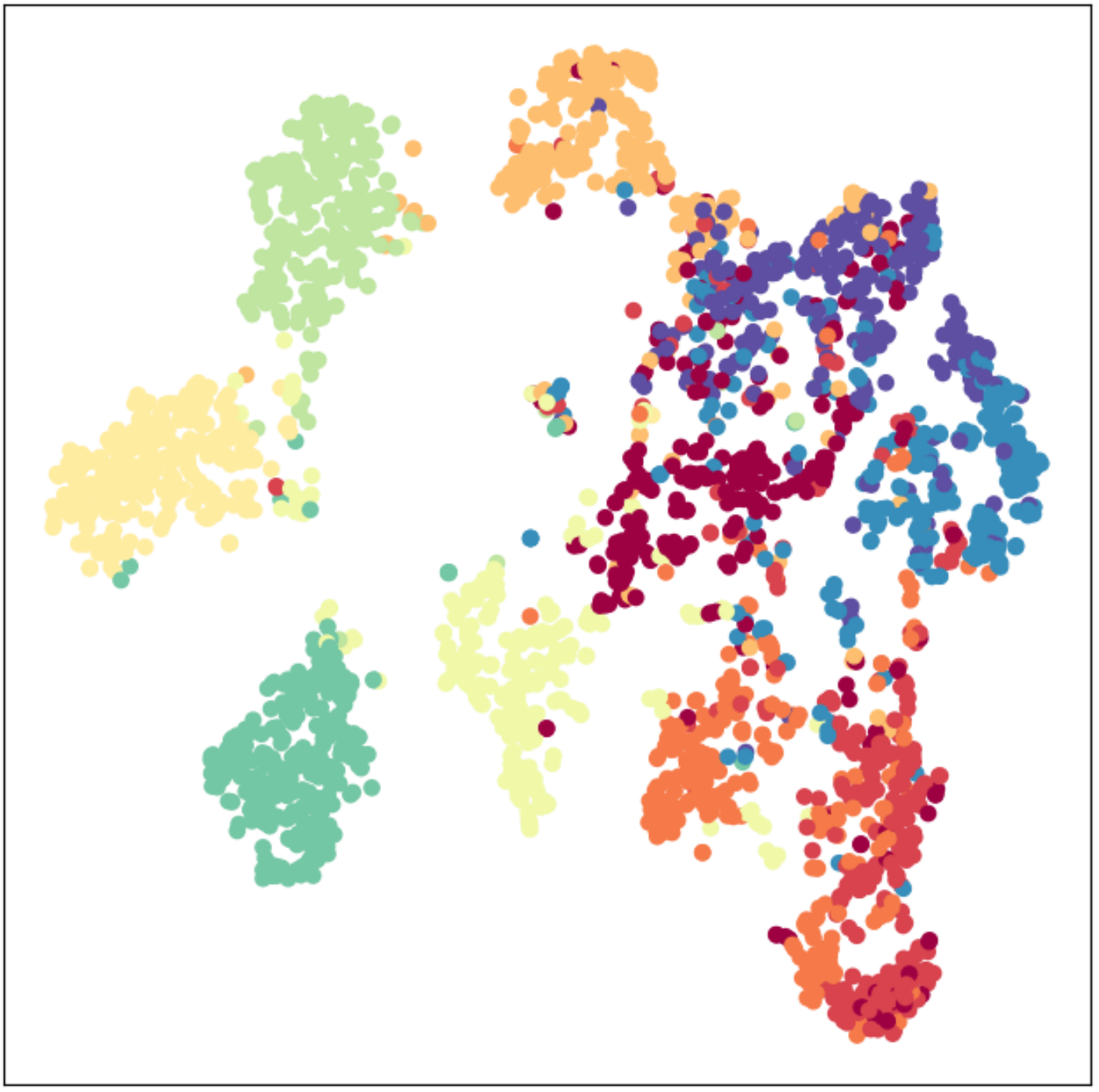}
    \label{fig_tsne_b3}}
    \hfil
    \subfloat[On PSP Learning w/o joint- and part-level contrast loss.]{\includegraphics[scale=0.3]{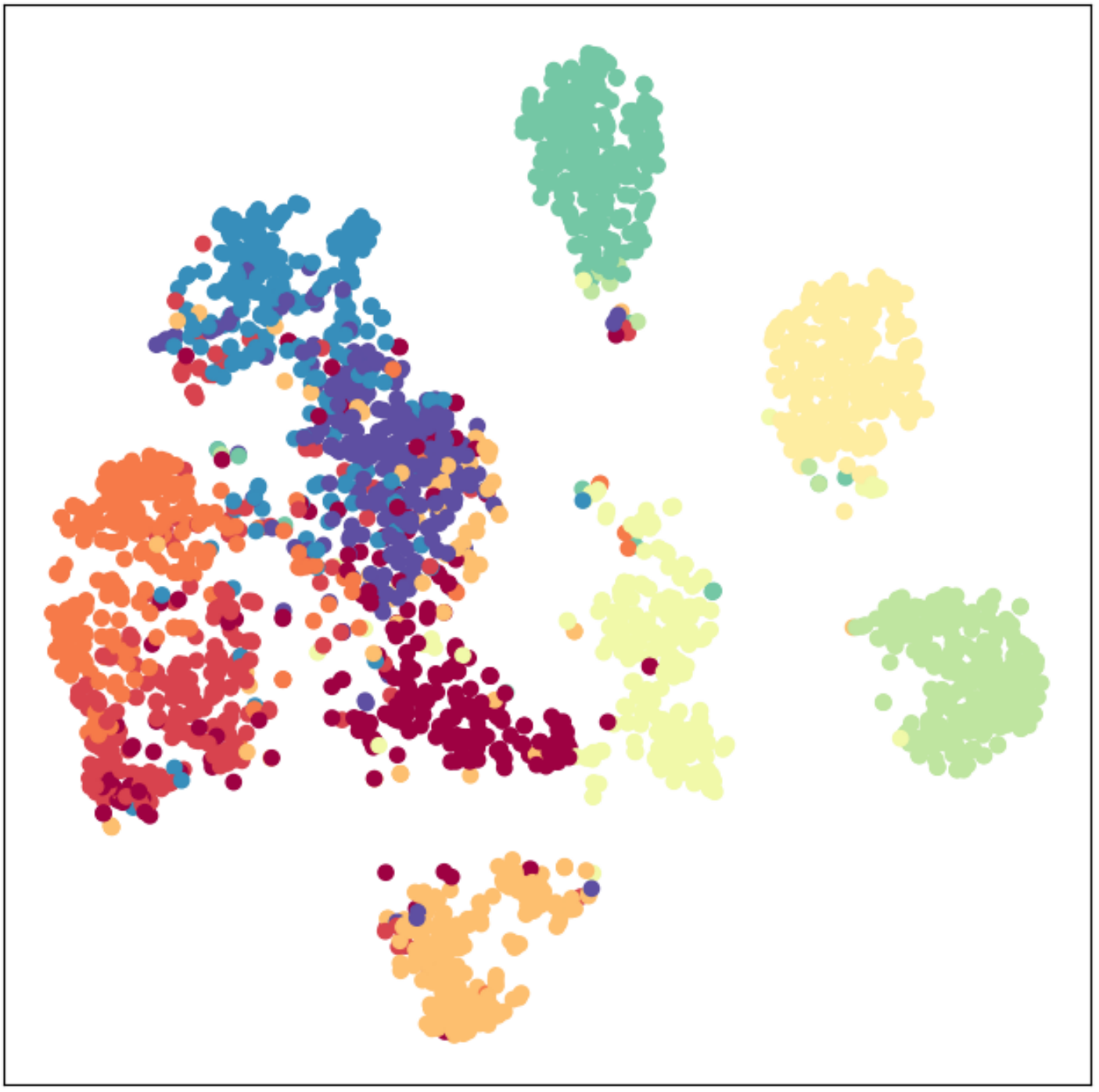}
    \label{fig_tsne_b4}}
    \\
    \subfloat[On PSP Learning w/o body-level contrast loss.]{\includegraphics[scale=0.3]{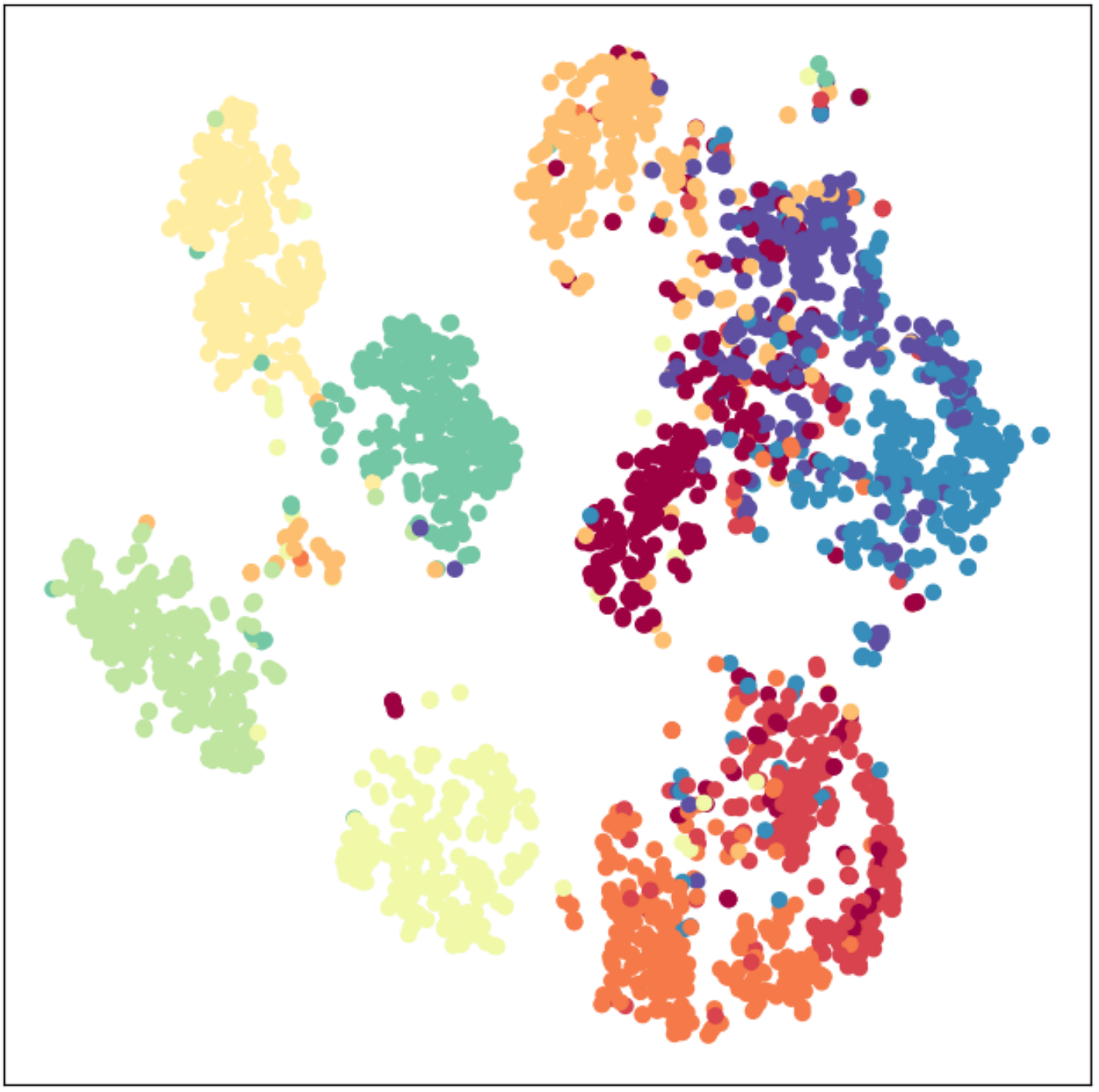}
    \label{fig_tsne_b5}}
    \hfil
    \subfloat[On PSP Learning w/o part-level contrast loss.]{\includegraphics[scale=0.3]{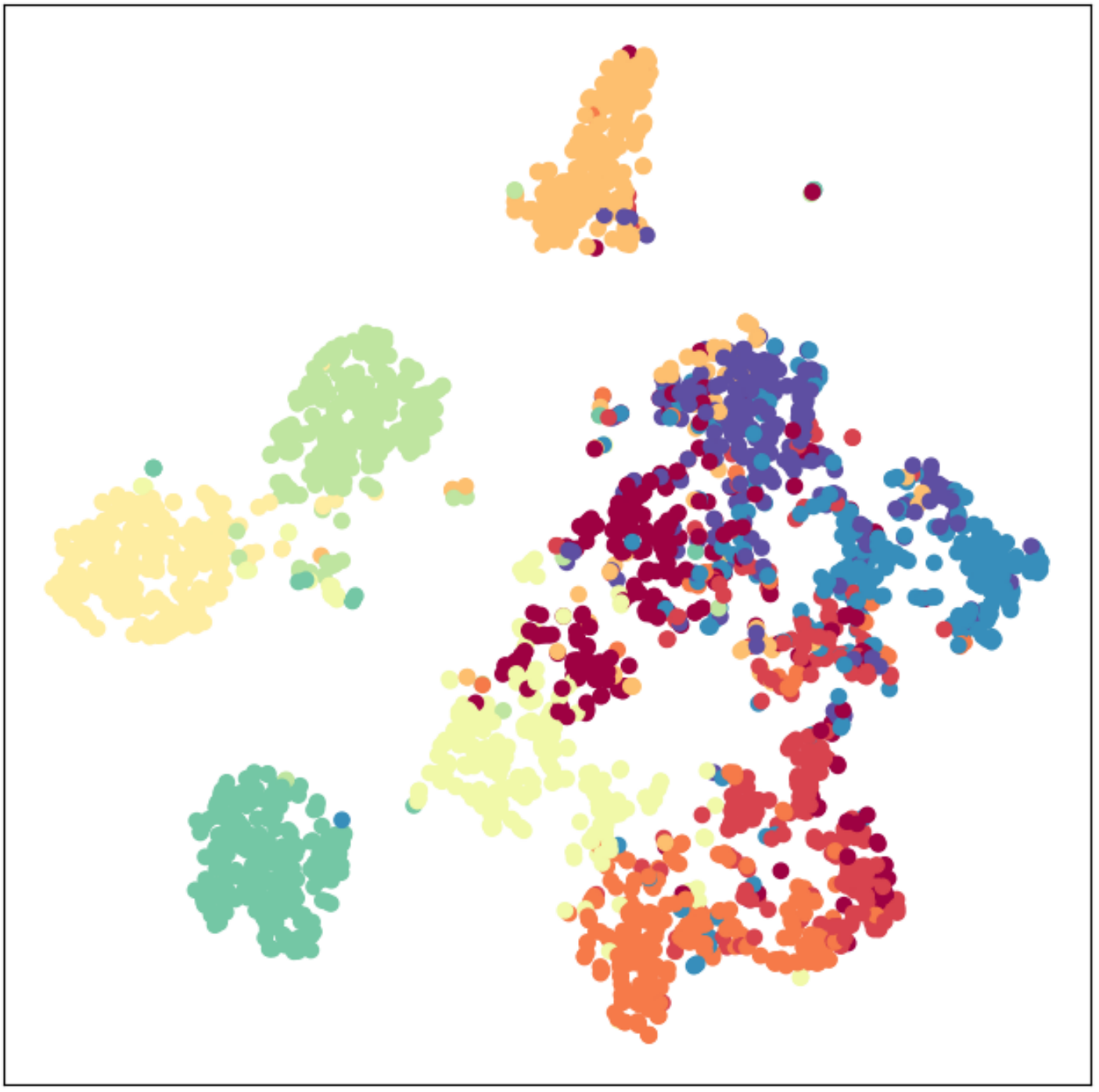}
    \label{fig_tsne_b6}}
    \hfil
    \subfloat[On PSP Learning w/o joint-level contrast loss.]{\includegraphics[scale=0.3]{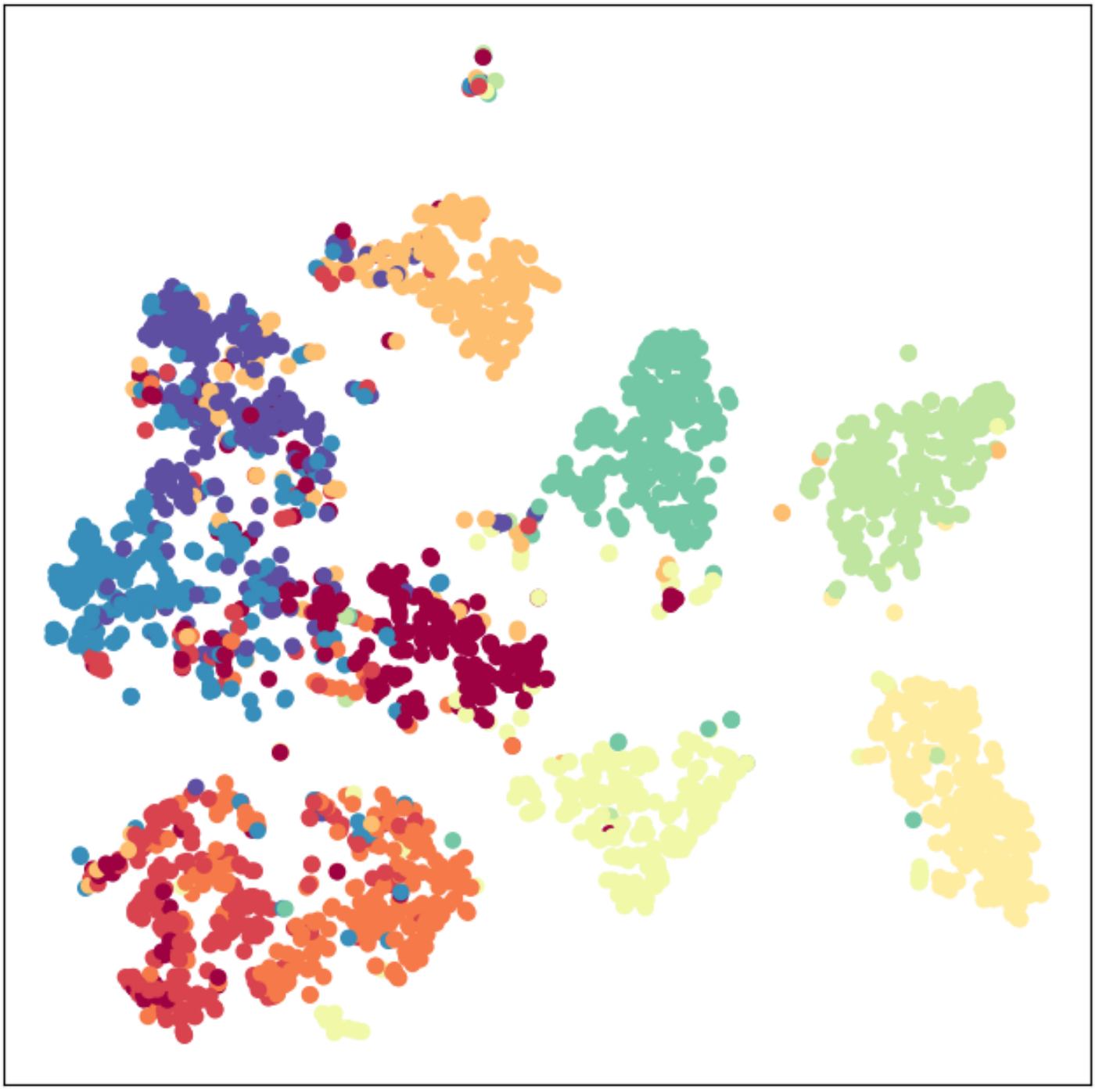}
    \label{fig_tsne_b7}}
    \hfil
    \subfloat[On PSP Learning.]{\includegraphics[scale=0.3]{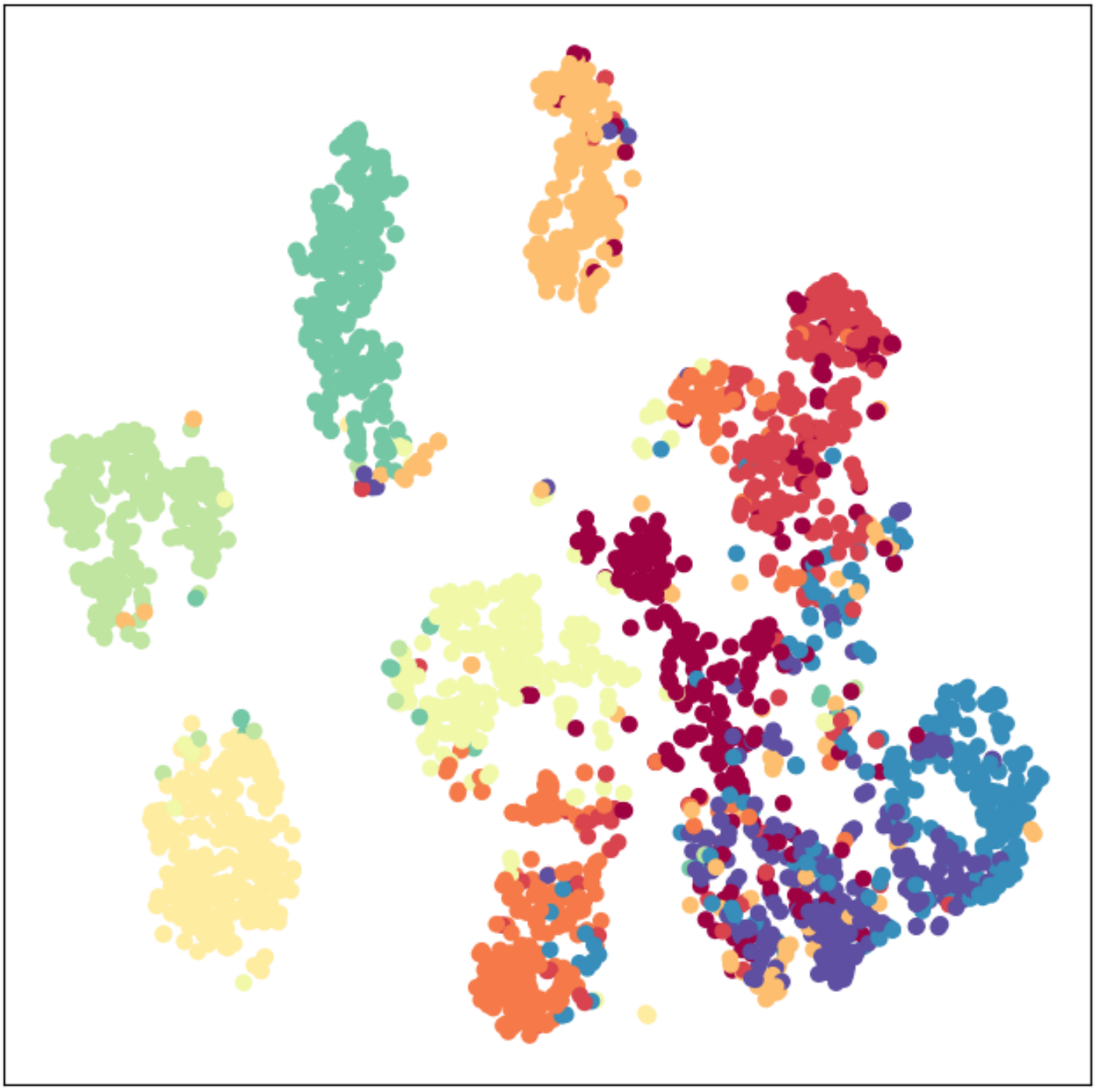}
    \label{fig_tsne_b8}}
    \caption{The t-SNE visualization of features (containing ten randomly selected action classes) obtained by different combinations in CCL on the NTU RGB+D dataset. Best view in the color PDF file.}
    \label{fig_tsne_contrast}
\end{figure*}

Moreover, to further explore the influence of various combinations of joint-level, part-level, and body-level contrast losses in CCL, we adopt t-SNE to visualize the distribution of features learned from PSP Learning under different baseline settings (e.g., B1, B2, $\cdots$, or B7). As can be seen from Figure~\ref{fig_tsne_contrast}, the combination of joint-level, part-level, and body-level contrast losses (namely CCL) can learn a more separable feature distribution than a single contrast loss to facilitate representation learning for action recognition. This is consistent with the quantitative results in Table~\ref{ablation}.

\section{Conclusion}
\label{conclusion}
In this work, we proposed a novel Pyramid Self-attention Polymerization Learning (PSP Learning) framework that complements the body-, part-, and joint-level semantic information to learn more comprehensive action representations, as the pretext task for addressing the problem of semi-supervised skeleton-based action recognition. In PSP Learning, there are two main insights, namely Pyramid Polymerizing Attention (PPA) and Coarse-to-fine Contrastive Loss (CCL). Specifically, PPA is capable to complement the body-, part-, and joint-level semantic information of skeleton actions by polymerizing the body-level attention map, part-level attention map, and joint-level attention map. CCL creatively measures the similarity of the body/part/joint-level contrasting features between joint and motion modality via body/part/joint-level contrast loss. For skeleton-based action recognition in the semi-supervised scenario, we comprehensively verify the effectiveness of the proposed PSP Learning by conducting extensive experiments on two public available datasets including NTU RGB+D and NW-UCLA. Considering the feasible portability of PPA and CCL, we will extensively explore the expandability of multi-attention transmission and multi-granularity contrast in the future.

\ifCLASSOPTIONcaptionsoff
  \newpage
\fi

\bibliographystyle{IEEEtran}
\bibliography{IEEEabrv,main}

\end{document}